\documentclass{article}

\usepackage[preprint]{neurips_2025}

\usepackage[utf8]{inputenc} 
\usepackage[T1]{fontenc}    
\usepackage{hyperref}       
\usepackage{url}            
\usepackage{booktabs}       
\usepackage{amsfonts}       
\usepackage{nicefrac}       
\usepackage{microtype}      
\usepackage{xcolor}         

\usepackage{graphicx}
\usepackage{subcaption}
\usepackage{wrapfig} 
\usepackage{ulem}
\usepackage{amsmath}

\title{QKCV Attention: Enhancing Time Series Forecasting with Static Categorical Embeddings for Both Lightweight and Pre-trained Foundation Models}

\author{%
  Hao Wang \\
  Independent Researcher \\
  Shanghai \\
  China \\
  \texttt{hw439@cornell.edu} \\
\And
  Baojun Ma \\
  Key Laboratory of Brain-Machine Intelligence for Information Behavior \\
  (Ministry of Education and Shanghai) \\
  School of Business and Management \\
  Shanghai International Studies University \\
  \texttt{mabaojun@shisu.edu.cn} \\
}

\begin{document}

\maketitle

\begin{abstract}
In real-world time series forecasting tasks, category information plays a pivotal role in capturing inherent data patterns. This paper introduces QKCV (Query-Key-Category-Value) attention, an extension of the traditional QKV framework that incorporates a static categorical embedding C to emphasize category-specific information. As a versatile plug-in module, QKCV enhances the forecasting accuracy of attention-based models (e.g., Vanilla Transformer, Informer, PatchTST, TFT) across diverse real-world datasets. Furthermore, QKCV demonstrates remarkable adaptability in fine-tuning univariate time series foundation model by solely updating the static embedding C while preserving pretrained weights, thereby reducing computational overhead and achieving superior fine-tuning performance.
\end{abstract}

\section{Introduction}

Time series prediction, which involves forecasting future values based on historical data arranged in chronological order, is a critical task in various domains such as finance, meteorology, transportation, and e-commerce.\cite{a001} Recently, significant advancements have been made in this field, particularly with the advent of deep learning technologies.

Deep Neural Networks (DNNs), known for their powerful nonlinear fitting capabilities and automatic feature extraction, have shown great potential in time series prediction. Convolutional Neural Networks (CNNs)\cite{a002} and Recurrent Neural Networks (RNNs)\cite{rumelhart1986learning, a1989learning} can capture complex patterns within time series data. \cite{guo2016learning,zeng2023financialtimeseriesforecasting,hochreiter1997long} However, DNNs also face challenges including high computational costs, training difficulties, and a lack of interpretability, demanding extensive further study.\cite{hussain2022design}

The integration of attention mechanisms into time series prediction has further elevated the field.\cite{huang2019dsanet} Attention mechanisms allow models to focus on important parts of the input data, providing more useful information and enhancing prediction accuracy. \cite{niu2021review} This has led to the development of advanced models like Transformer\cite{vaswani2017attention}, Temporal Fusion Transformers (TFT)\cite{lim2021temporal}, Hierarchical Transformer (HT)\cite{wang2024considering}, and PatchTST\cite{nie2022time} which have demonstrated state-of-the-art performance in various time series prediction tasks. Today, attention mechanisms and transformers form the foundation for many cutting-edge AI artifacts, especially large language models (LLMs) for NLP\cite{ouyang2022training,touvron2023llama,du2021glm} and time series forecasting as well, with notable examples including the TimeGPT\cite{liao2025timegpt}, GPHT\cite{liu2024generative} and TimeFM\cite{das2024decoder}. 

In time series analysis and prediction practices, the significance of categorical information cannot be overlooked, as most time series contain categorical covariates with observations recorded at each point in time. \cite{tymchuk2022forecasting} Categorical covariates, such as product catalogs, geographical attributes (e.g., country, region), and demographic data (e.g., age, income), can exert a substantial influence on time series patterns, enabling better capture of demand trends and seasonal variations. For instance, in the agriculture industry, the demand for watermelon and other summer fruits tends to peak during the warmer months, and often aligns with the harvesting season for these fruits, which are sensitive to temperature and weather conditions. In contrast, the demand for root vegetables like carrots and potatoes peaks in cooler seasons, as they are more resilient to colder weather and can be stored for longer periods. Meanwhile, staple crops like wheat and rice maintain a relatively stable demand throughout the year, serving as essential components of global food supply chains and less influenced by short-term weather fluctuations.\cite{vanitha2021growth} Integrating these categorical features into prediction algorithms for agriculture industry enables the model to recognize and differentiate between various trend patterns for each individual product, ultimately boosting its robustness and prediction accuracy.

Today, models incorporating exogenous variables typically treat categorical covariates as static features, processing them alongside dynamic features. \cite{oreshkin2019n,challu2023nhits,lim2021temporal} However, little research has explored integrating such covariates directly into the attention mechanism. Furthermore, pre-trained foundational models are often trained in a univariate (or channel-independent) manner, limiting their adaptability when real-world users fine-tune them with additional static features. In this work, we systematically investigate this gap by exploring novel dot-product attention mechanisms for time series forecasting. Our key contributions are as follows:

\begin{itemize}
\item We propose a novel approach, QKCV attention for directly incorporating categorical information into the attention layer of time series forecasting models, along with three variations to implement this approach. This enables the model to better capture the influence of categorical features on temporal patterns.
\item To demonstrate its applicability across attention-based models, we have modified several existing time series forecasting frameworks, including the Vanilla Transformer, Informer, PatchTST, and TFT, to incorporate these variations. Our experiments confirm that the method is compatible with attention-based models.
\item We integrate this mechanism into TimeFM\cite{das2024decoder}, an open-source pretrained time-series foundation model developed by Google Research. Empirical results demonstrate that our approach enables efficient integration of static features into univariate pretrained models, achieving lower memory overhead and consistent performance gains.
\end{itemize}

\section{Related work}

In the realm of time series prediction with Deep Neural Networks (DNNs), contributions centered around Recurrent Neural Networks (RNNs)\cite{a1989learning} and its derivatives including Gated Recurrent Units (GRUs) \cite{salem2022gated}, Long Short-Term Memory (LSTM)\cite{graves2012long}, DeepAR\cite{salinas2020deepar}, Neural Basis Expansion Analysis for Time Series Forecasting(N-BEATS)\cite{oreshkin2019n}, and Neural Hierarchical Interpolation for Time Series Forecasting(NHITS)\cite{challu2023nhits}.

Attention mechanisms have emerged as pivotal components in the realm of time series forecasting. The self-attention mechanism is a process whereby a query and a sequence of key-value pairs are mapped to produce a corresponding output vector. The output attention vector is determined by the weighted summation of values V, with the weights being computed based on the relationship between the query Q and the corresponding key K, as deplicited in formula below:

\begin{equation}
  Attention(Q, K, V) = {softmax}\left(\frac{QK^T}{\sqrt{d_k}}\right)V
\end{equation}

The formula uses the transpose of K, denoted as $K^T$, for multiplication with Q. $d_k$ denotes the dimension of the query and key. Before normalized using the softmax function, the score matrix is divided by a scaling factor of $\sqrt{d_k}$. 

Among the most notable advances of attention mechanisms is the introduction of the Transformer architecture, which was initially designed for sequence-to-sequence tasks in NLP.\cite{vaswani2017attention} The Transformer leverages multi-head self-attention mechanisms to capture both short-term and long-term dependencies through pairwise (query-key) interactions\cite{han2021transformer}, and has been adapted to time series forecasting with various optimizations including Informer\cite{zhou2021informer}, LogSparse Transformer\cite{li2019enhancing}, TFT\cite{lim2021temporal}, PatchTST\cite{nie2022time}, and Fredformer\cite{piao2024fredformer}. These models collectively represent state-of-the-art approaches for time series forecasting, each addressing unique challenges. 

Nowadays, a plethora of large-scale models have emerged in the field of forecasting. Notable examples include TimeGPT\cite{liao2025timegpt}, TimeFM\cite{das2024decoder}, AutoGluon\cite{shchur2023autogluon}, GPHT\cite{liu2024generative}, and TabPFN\cite{hollmann2022tabpfn}, all of which are designed to undergo pre-training on large-scale time-series data to enhance their generalization capabilities. On the other hand, to adapt large-scale models to specific downstream tasks or enhance their performance in resource-constrained environments, numerous finetuning methods have been proposed, such as LoRA\cite{devalal2018lora}, QLoRA\cite{dettmers2023qlora}, VeRA\cite{kopiczko2023vera}, AdapterFusion\cite{bhardwaj2024rapid}, and so on.

\section{Methodology}
\subsection{Model intuition}
\begin{wrapfigure}{r}{0.5\textwidth}  
\vspace{-0.5em}
  \centering
  \includegraphics[width=\linewidth]{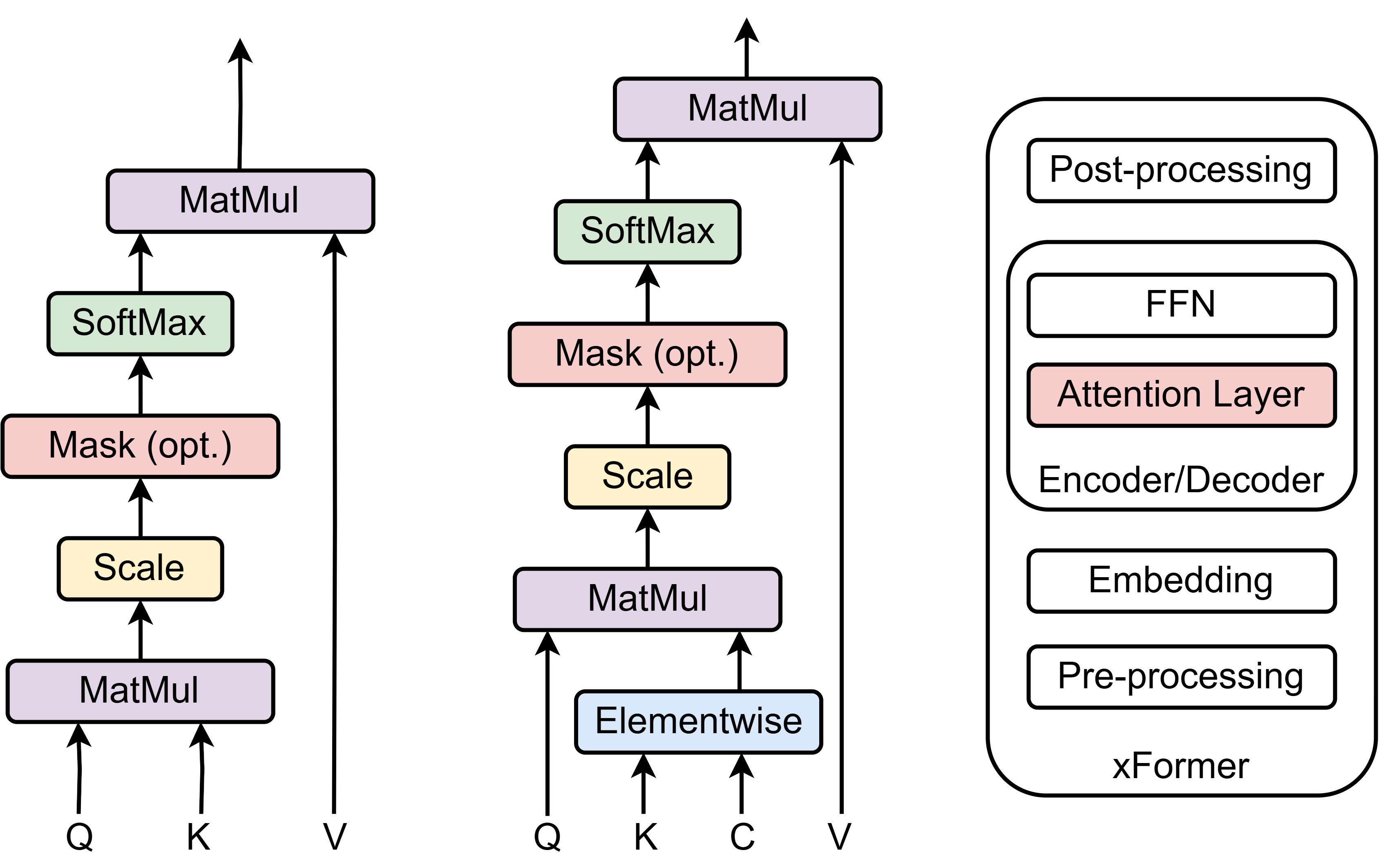}
  \caption{A comparison between the Dot Product Attention mechanism (left) and our proposed modified QKCV attention mechanism incorporating static categorical embeddings (middle). The figure on the right shows a typical structure of a Transformer variant and the location of the attention module (in red) that we modified.}
  \label{fig:DotProductAttention}
\vspace{-0.5em}
\end{wrapfigure}
Patterns in time series may be influenced by categorical covariates, as products belonging to the same or similar categories, families, or clusters often exhibit similarities in their time-varying patterns. This phenomenon is prevalent in practical scenarios, such as when specific types of products experience peak popularity during certain times of the year, whereas others maintain stable sales throughout. Furthermore, distinct trends can be discerned based on the nature of the products.

In previous methodologies, categorical information was typically processed alongside time-varying feature series and subsequently input into encoders primarily for feature extraction purposes. However, within the self-attention layer, the query-key matching mechanism may not adequately incorporate the contextual information provided by categorical features. This oversight can hinder the self-attention module's ability to distinguish between observed values that belong to a characteristic pattern of a particular category, a change point, or an anomaly. Consequently, this may give rise to underlying interpretability challenges.

Another challenge arises with pre-trained foundation models. These models are typically trained on univariate data, requiring both the target sequence and additional covariates to be flattened into a single-dimensional input format. However, since the model inherently treats all inputs and outputs as predictable sequences, forcibly integrating supplementary information into these sequences can distort the data distribution and semantic meaning, potentially degrading prediction accuracy.

\begin{figure}[htbp]
    \centering
    \begin{subfigure}[t]{0.54\textwidth}
        \centering
        \includegraphics[width=\textwidth]{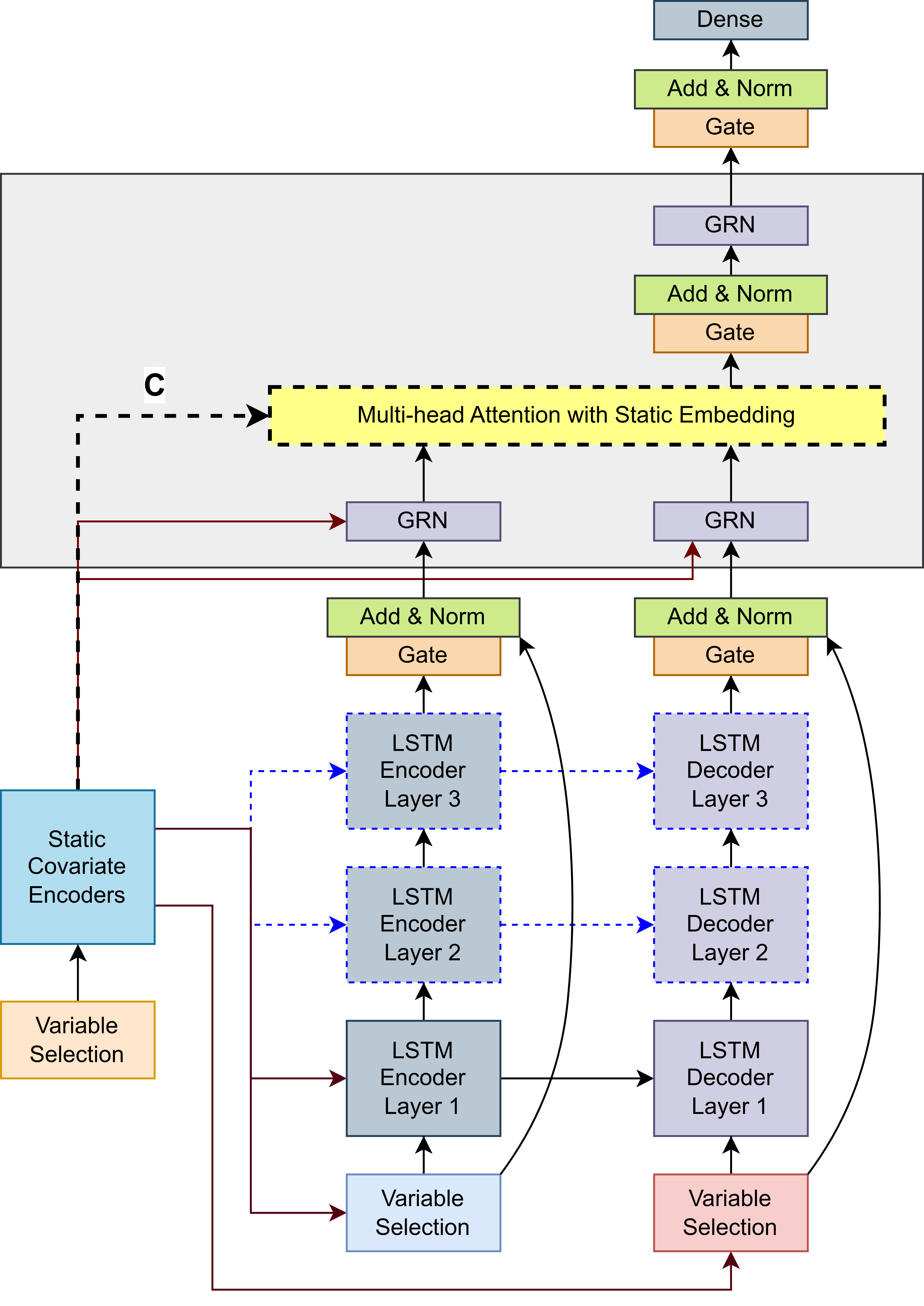}
        \caption{TFT with QKCV}
        \label{figsubTransformerWithStaticFeatures0}
    \end{subfigure}
    \hfill
    \begin{subfigure}[t]{0.44\textwidth}
        \centering
        \includegraphics[width=\textwidth]{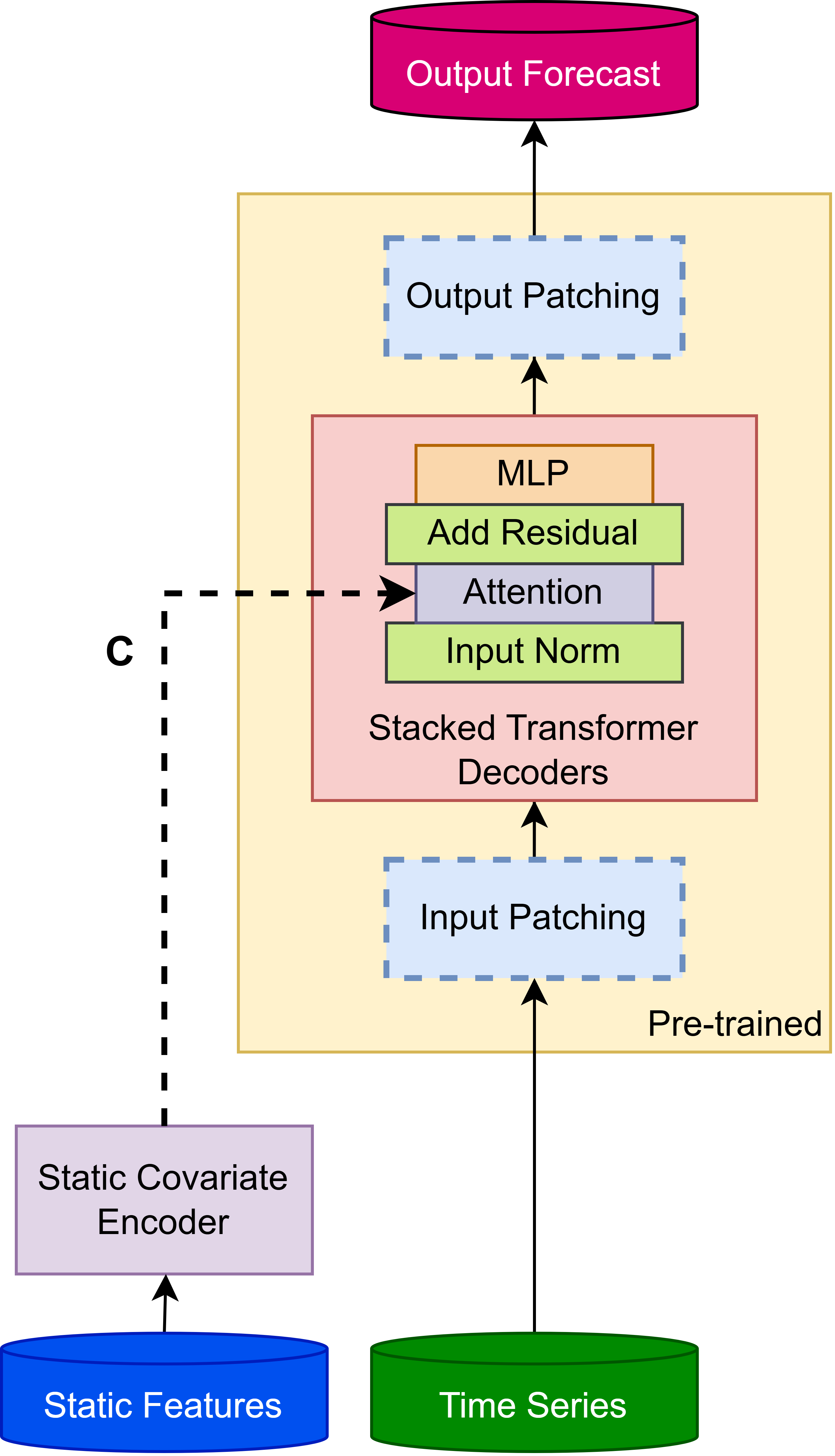}
        \caption{TimeFM with QKCV}
        \label{figsubTransformerWithStaticFeatures1}
    \end{subfigure}
    \caption{Our modified TFT structure (a) and TimeFM (b) serves as examples of applying the QKCV mechanism to attention-based lightweight model and pre-trained fundation model, where static embedding vector C, derived from Static Covariates Encoders, is directly fed into the Multi-head Attention layer, as indicated by the black dashed line. The code of both models and others are available in the GitHub repository.}
    \label{fig:TransformerWithStaticFeatures}
\vspace{-0.5em}
\end{figure}

To overcome these limitations, we propose a novel approach termed the QKCV (Query-Key-Category-Value) attention, that effectively integrates auxiliary features while preserving the original sequence semantics. Our method is designed to maintain full compatibility with existing model architectures and minimize interference with pre-trained weights, ensuring stable fine-tuning performance.

Specifically, we introduce direct injection of categorical embeddings into the attention layer. These embedding vectors are generated from the static categorical attributes of each entity, creating feature-specific representations optimized for attention mechanisms. We implement and evaluate this approach with Dot-Product Attention, as it represents the most widely adopted attention scheme in self-attention based models.

\subsection{Formulation}

We introduce categorical embeddings, denoted as ${C \in R ^{N\times d_c}}$, into the self-attention mechanism to enhance its ability to model categorical relationships within sequences. The attention computation is modified as below:

\begin{equation}
  {Attention}(Q, K, V) = {softmax}\left(\frac{Q \left(K \odot \mathbf{C}\right)^T}{\sqrt{d_k}}\right) V
\end{equation}

where $\odot$ denotes a learned element-wise operation, as illustrated in Figure 1. The modified attention score $\alpha$ between query q and key k becomes:
\begin{equation}
\alpha_{ij} = \frac{\exp\left(q_i \cdot (k_j \odot c_j)^T / \sqrt{d_k}\right)}{\sum_{m=1}^N \exp\left(q_i \cdot (k_m \odot c_m)^T / \sqrt{d_k}\right)},
\end{equation}
where ${c \in R ^{d_c}}$ is the categorical embedding for category j. This formulation ensures that keys from different categories contribute differently to the attention distribution, depending on their compatibility with the query and the category-specific embedding.

\subsection{Implementations}
\paragraph{\textbf{Static feature embedding}} 
During training, the input static exogenous variable tensor V possesses a size of $[B\times F]$, where B denotes the batch size and F represents the count of static features. After first-step embedding, V is transformed into V' with a size of $[B\times E]$, where E signifies the length of each embedding vector. It is important to note that the matrix K has a size of $[B\times L\times H\times D]$, where L indicates the length of the time series, H stands for the number of attention heads, and D represents the dimension of each head. To facilitate the calculation, we expand V' for L times (as static information remains unchanged over time), and set $E=H\times D$, to get a vector with dimensions $[B\times E\times L]$. Then we can appropriately reshape the vector to acquire C with a size of $[B\times L\times H\times D]$, which is congruent with the shape of K.

For the embedding step of transforming V into V', there are numerous approaches like Multi-Layer Perceptrons (MLPs). Specifically, in this research, we employ TFT's Static Covariate Encoder (SCE) to embed categorical features, given its demonstrated ability to select significant variables and eliminate unnecessary noisy inputs. The encoder comprises a Variable Selection Network module followed by a GRN layer, which is described in detail in  \cite{lim2021temporal}.

\paragraph{\textbf{Attention variants}} 

We introduce three variants of the $\odot$ operation to integrate categorical information, dynamically adjusting attention score magnitudes while maintaining the semantic consistency of parameters.

\textbf{QKCV-v1: Gated residual network (GRN)-based feature fusion.} 
In this approach, the static categorical feature embedding C is first passed through a Gated Residual Network (GRN) layer to dynamically modulate its representation. The resulting GRN(C) is then element-wise multiplied with the key matrix K:
\begin{equation}
\text{QKCV-v1} = {softmax}\left(\frac{Q\left(K \times {GRN}(C)\right)^T}{\sqrt{d_k}}\right)V
\end{equation}
The element-wise multiplication ($\times$) allows the GRN-enhanced categorical features to selectively emphasize or suppress elements of K, thereby adapting the attention mechanism to the specific characteristics of the input data. This design choice is grounded in the observation that categorical features often exhibit non-linear interactions with query-key dynamics, which can be effectively captured by a learnable gating mechanism (as implemented by GRN). Besides, the introduction of C does not impact the generation of K, which is typically derived from a linear transformation of input features, thereby preserving the original Dot Attention mechanism intact.

\textbf{QKCV-v2: Probabilistic scaling via sigmoid activation.}
Recognizing that the K$\odot$C operation can be interpreted as scaling each element of K by a learned probability, we introduce a sigmoid activation function to constrain the scaled values within the range [0,1]:
\begin{equation}
\text{QKCV-v2} = {softmax}\left(\frac{Q\left(K \times {Sigmoid}({GRN}(C))\right)^T}{\sqrt{d_k}}\right)V
\end{equation}
The sigmoid function ensures that the categorical features act as probabilistic gates, modulating the influence of K in a manner that is both interpretable and stable. This variant is particularly suitable for scenarios where categorical information is expected to have a bounded impact on the attention scores, such as in multi-modal fusion tasks where categorical features should not dominate the attention mechanism.

\textbf{QKCV-v3: Residual connection-inspired feature integration.}
Inspired by the success of residual connections in deep learning, we propose a variant that establishes a direct pathway between the static categorical embedding C and the attention mechanism:
\begin{equation}
\text{QKCV-v3} = {softmax}\left(\frac{Q\left(K + {GRN}(C)\right)^T}{\sqrt{2 \cdot d_k}}\right)V
\end{equation}

Here, $+$ denotes element-wise addition, and the divisor $\sqrt{2 \cdot d_k}$ accounts for the increased scale of the dot product due to the summation of two embedding vectors. This variant is motivated by the hypothesis that categorical features can provide complementary information to the query-key interactions, and that a residual connection can facilitate the flow of this information without disrupting the underlying attention dynamics. Empirically, this approach has been shown to improve the robustness of the attention mechanism, particularly in low-resource settings where categorical features may be noisy or sparse.

To evaluate the performance of these mechanisms on time-series forecasting datasets, we incorporate the aforementioned modifications into the Vanilla Transformer, Informer, patchTST, and TFT models. How their attention modules modified is demonstrated in Figure~\ref{fig:DotProductAttention}, and to be more specific, Figure~\ref{fig:TransformerWithStaticFeatures} illustrates the architectural overview of our modified TFT and TimeFM models.

\section{Experiments}

\subsection{Real-world datasets}
\begin{wraptable}{r}{0.52\linewidth}
\centering
\caption{Statistics of experimental datasets}
\label{tab:modelsconf}
\begin{tabular}{llll}
\toprule
	Dataset & {Meal} & {Favorita} & {M5}  \\
\midrule
\textbf{Dataset details} &   &   &    \\
Target type      &int   &int & int  \\ 
Freq &weekly   &daily & daily  \\ 
Forecasting horizon&10&30   &28   \\ 
Number of entities &3.6k&143k   &  30k      \\
Number of samples  & 402k&25m   & 59m  \\
\bottomrule
\vspace{-0.5em}
\end{tabular}
\end{wraptable}
To evaluate the effectiveness of our proposed improvements, we require real-world datasets that incorporate meaningful categorical features. While such data is commonly available within enterprise settings, identifying suitable publicly accessible datasets remains challenging. Existing popular benchmarks for time series forecasting (e.g., ETT, ECL, Exchange, Traffic, and Weather) are either synthetic or lack sufficient covariates. To address this limitation, we employ three real-world datasets—Meal, Favorita, and M5—to assess our enhancements. The statistical properties of these datasets are summarized in Table~\ref{tab:modelsconf}, with further details provided in the Supplementary Materials.

\begin{table*}[htbp]
\centering
\small
  \caption{Performance metrics (WPE, P50, P90) of lightweight models on real-world datasets. The best performance is underlined with \uuline{double line} and the second best with \uline{single line}.}
\centering
	\label{tab:modelsperformance}
\begin{tabular}{p{2.1cm} p{0.8cm} p{0.8cm} p{0.8cm}|p{0.8cm}p{0.8cm}p{0.8cm}|p{0.8cm}p{0.8cm} p{0.8cm} }
	\toprule
	Model & {} & {WPE} & {} & {} & {P50} & {} & {} & {P90} & {} \\
  \cmidrule{2-10} 
	{} & {Meal}& {Favorita} & {M5}   & {Meal}&{Favorita} & {M5}  & {Meal}&{Favorita} & {M5} \\
\midrule
DeepAR   &0.2958& 0.5015&0.7143$^{\star}$&0.3971&0.6038 &0.8157$^{\star}$&0.2632&0.4431&0.7038$^{\star}$\\
NBEATSx  &0.2058&0.2674&0.7591$^{\star}$&0.3539&0.4838 &0.9520$^{\star}$&0.2431&0.3250&0.7602$^{\star}$\\ 
NHITS      &{0.1989} &0.2572&0.7590$^{\star}$ &0.3343&0.4758 &0.9519$^{\star}$&0.2552&0.3291  &0.7622$^{\star}$  \\
\cmidrule{1-10}
Transformer &0.1992 &0.2198&0.4405 &0.3256&0.4810 &0.6935&0.2322&0.3272&0.6416  \\
with QKCV-v1 &0.1978 &\uline{0.2134}&0.3652 &0.3243&0.4783 &0.6901&0.2310&0.3246&0.5448  \\
with QKCV-v2 &0.1999 &0.2165&0.4076 &0.3247&0.4785 &0.6891&0.2337&0.3261&0.5767  \\
with QKCV-v3 &0.1987 &\uuline{0.2123}&0.4530 &0.3249&0.4733 &0.6918&0.2306&0.3184&0.6489  \\
\cmidrule{1-10}
Informer &0.2062 &0.2406 &0.3978 &0.3324 &0.4764 &0.6712 &0.2427 &0.3249 &0.5477  \\
with QKCV-v1 &0.2061 &0.2255 &0.3962 &0.3324 &0.4730 &0.6706 &0.2434 &0.3187 &0.5571  \\
with QKCV-v2 &0.2058 &0.2358 &0.3775 &0.3320 &0.4756 &0.6706 &0.2422 &0.3209 &0.5419  \\
with QKCV-v3 &0.2063 &0.2303 &0.3852 &0.3321 &0.4770 &0.6714 &0.2421 &0.3227 &0.5534  \\
\cmidrule{1-10}
PatchTST &0.1982 &0.2859 &0.3982 &0.3430 &0.5027 &0.6919 &0.2662 &0.3222 &0.5860  \\
with QKCV-v1 &\uline{0.1971} &0.2764 &0.3645 &0.3426 &0.5009 &0.6928 &0.2666 &0.3393 &0.5607  \\
with QKCV-v2 &0.1982 &0.2709 &0.3791 &0.3427 &0.4923 &0.6912 &0.2671 &0.3225 &0.5584  \\
with QKCV-v3 &\uuline{0.1946} &0.2514 &0.3619 &0.3414 &0.4728 &0.6948 &0.2624 &0.3068 &0.5685  \\
\cmidrule{1-10}
TFT          &0.2010 &{0.2237}&0.3686 &0.3222&0.4753 &0.6585&0.2265&0.3176 &0.5217  \\
with QKCV-v1   &{0.1986} &{0.2201}  &0.3659    &\uline{0.3216}&\uuline{0.4627} &\uuline{0.6568}  &\uline{0.2230}&0.3123&\uline{0.5187} \\
with QKCV-v2   &0.2044&0.2321  &\uuline{0.3515}    &0.3243&\uline{0.4632}&0.6580  &0.2252&\uuline{0.3087}&0.5246 \\
with QKCV-v3   &0.1991&0.2262  &\uline{0.3536}&\uuline{0.3198}&0.4652   &\uline{0.6571}&\uuline{0.2140}&\uline{0.3100}&\uuline{0.5144}\\
\bottomrule
\vspace{-0.5em}
\end{tabular}
\end{table*}

\subsection{Training}
\begin{wraptable}{r}{0.65\linewidth}
\vspace{-1.5em}
\centering
\caption{GPU memory utilization (GB) across fine-tuning strategies of modified TimeFM.}
\label{tab:gpumemusage}
\begin{tabular}{llll}
\toprule
	Dataset & {Meal} & {Favorita} & {M5}  \\
\midrule
Patching layer optimization (PL) &3.79   &6.18 &7.61  \\ 
PL + QKCV                       &4.97   &6.47 & 8.21  \\ 
Full parameter fine-tuning (FP) &11.19&13.59   &15.66   \\ 
FP + QKCV                    &12.13&14.14& 15.80      \\
\bottomrule
\vspace{-1.5em}
\end{tabular}
\end{wraptable}
For each dataset, we partition all time series into three distinct subsets: training, validation, and test sets. Hyperparameter optimization for lightweight models is performed on the validation set using the Tree-structured Parzen Estimator (TPE) search algorithm\cite{bergstra2011algorithms}. We conduct 50 TPE iterations per dataset, with complete hyperparameter search ranges provided in the Supplementary Materials. The optimized hyperparameters are then applied to the test set for final performance evaluation. To ensure fair comparison, we maintain identical hyperparameters across both baseline models and our modified architectures, which integrate the three proposed QKCV attention variants (v1-v3). For TimeFM, we address the mismatch between the original patch length and forecast horizons by adapting the patching layers to each dataset's temporal characteristics, co-training them with static covariate encoders . This adapted configuration, designated as "Patching Layer Optimization (PL)", serves as the baseline implementation and is documented in Table~\ref{tab:modelsperffm}, while its architectural representation appears as light blue rectangles in Figure~\ref{figsubTransformerWithStaticFeatures1}. All TimeFM experiments employ a uniform training configuration with maximum iterations set to 1,000 steps and a batch size of 256 samples per update.

We use Python libray NeuralForecast v1.7.5, developed by Nixtla\cite{olivares2022libraryneuralforecast} for lightweight models, and TimeFM\cite{das2024decoder} as pre-trained large model. Our code and training configurations have been published on GitHub and will be available upon acceptance.
\begin{table*}[htbp]
\centering
  \caption{WPE and MAE losses across datasets for different fine-tuning strategies of modified TimeFM. The best performance is underlined with \uuline{double line} and the second best with \uline{single line}.}
\centering
	\label{tab:modelsperffm}
\begin{tabular}{lll|ll|ll}
	\toprule
	Dataset & {Meal}  & {} & {Favorita} & {} & {M5} & {} \\
  \cmidrule{2-7} 
	Metric & {WPE} & {MAE}  & {WPE}& {MAE} & {WPE}& {MAE}  \\
\midrule
Patching layer optimization (PL)  & 0.3518 & 111.3217 & 0.2683 & 3.6552 & 0.2855 & 1.0831  \\
PL + QKCV-v1  & 0.3347 & 109.8986 & 0.2619 & 3.6368 & 0.2798 & \uuline{1.0698} \\
PL + QKCV-v2  & 0.3153 & \uuline{103.8176} & 0.2679 & \uline{3.6229} & \uline{0.2778} & 1.0752 \\
PL + QKCV-v3  & \uline{0.3120} & \uline{104.9749} & 0.2586 & \uuline{3.5774} & 0.2798 & \uline{1.0716} \\
\cmidrule{1-7}
\textbf{Ablation studies}                       &  &  & & & & \\
Full parameter fine-tuning (FP) & 0.4193 & 133.7287 & \uline{0.2562} & 3.6333 & 0.3958 & 1.4553 \\
FP + QKCV-v1  & 0.4077 & 124.2514 & \uuline{0.2536} & 3.7469 & \uuline{0.2749} & 1.0837 \\
FP + QKCV-v2  & \uuline{0.3096} & 118.0344 & 0.2641 & 3.8410 & 0.2891 & 1.1386 \\
FP + QKCV-v3  & 0.3286 & 120.1289 & 0.2908 & 4.0417 & 0.2782 & 1.0940 \\
PL + QKCV-v1(MLP encoder)  & 0.3222 & 107.5909 & 0.4831 & 8.3584 & 0.4003 & 2.7082  \\
PL + QKCV-v2(MLP encoder)  & 0.3326 & 111.1009 & 0.3439 & 4.2444 & 0.2882 & 1.0935 \\
PL + QKCV-v3(MLP encoder)  & 0.3419 & 110.2009 & 0.2935 & 4.0341 & 0.3410 & 1.2727 \\
PL + MLP as input compressor    & 0.8249 & 188.3073 & 0.9493 & 6.9375 & 0.9997 & 1.8018 \\
PL + SCE as input compressor    & 0.8395 & 190.2695 & 0.8205 & 6.1941 & 0.9830 & 1.7793 \\
\bottomrule
\end{tabular}
\end{table*}

\subsection{Results and performance}

For lightweight models, we evaluate our approach by comparing it with the vanilla Transformer, Informer, patchTST, and TFT, as well as other state-of-the-art RNN-based time series forecasting models such as DeepAR, NBEATSx, and NHITS. Among these models, TFT, DeepAR, NBEATSx, and NHITS have the capability to incorporate static categorical features as input, which makes them suitable for a comprehensive comparison. In contrast, the vanilla Transformer, Informer, and patchTST do not leverage exogenous categorical variables. The purpose of mentioning this is to demonstrate that the new mechanism can serve as a general solution for introducing categorical information into all attention-based models, enhancing their accuracy.
\begin{figure}[htbp]
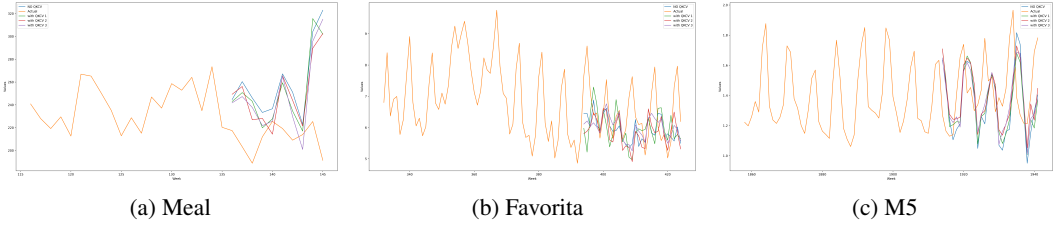

    \centering
    \begin{subfigure}[t]{0.32\textwidth}
        \centering
        \includegraphics[width=\textwidth]{images/Meal\_20250512\_QKCV.png}
        \caption{Meal}
        \label{figavgforecast:figsub0}
    \end{subfigure}
    \hfill
    \begin{subfigure}[t]{0.32\textwidth}
        \centering
        \includegraphics[width=\textwidth]{images/Favorita\_20250512\_QKCV.png}
        \caption{Favorita}
        \label{figavgforecast:figsub1}
    \end{subfigure}
    \hfill
    \begin{subfigure}[t]{0.32\textwidth}
        \centering
        \includegraphics[width=\textwidth]{images/M5\_20250512\_QKCV.png}
        \caption{M5}
        \label{figavgforecast:figsub2}
    \end{subfigure}
    \caption{Averaged real values versus forecast outputs from TimeFM for each dataset.}
    \label{figavgforecast}
\end{figure}

To assess the performance of our mechanism, we calculate several evaluation metrics, including the Weighted Percentage Error (WPE), and the Weighted Quantile Loss at the 50th percentile (P50) and 90th percentile (P90). Table~\ref{tab:modelsperformance} presents the results. The comparision among each cluster (i.e., PatchTST with QKCV vs. vanilla PatchTST) demonstrates that our modification of the QKCV attention mechanism has significantly reduced both WPE and the P50/P90 quantile losses, except in some rare cases. Notably, the TFT model with our proposed modifications outperforms not only the vanilla TFT baseline but also other state-of-the-art forecasting models, highlighting the effectiveness of our approach. Besides, the RNN-based models appear to perform poorly on the extensive M5 dataset, despite using even more iterations of TPE optimization (specifically 80 iterations, compared to only 50 for TFT) and larger hyperparameter ranges, still achieved much poorer performance, as indicated with ${\star}$ in Table~\ref{tab:modelsperformance}.

For the pre-trained model TimeFM, we evaluate performance using Weighted Percentage Error (WPE) and Mean Absolute Error (MAE) metrics, as shown in Table~\ref{tab:modelsperffm}. The results demonstrate that QKCV-enhanced models consistently outperform the original architecture across most evaluation scenarios. Figure~\ref{figavgforecast} compares the averaged ground truth values with corresponding forecast outputs across all experimental datasets. The temporal trajectories demonstrate that our QKCV-enhanced model achieves closer alignment with actual observed patterns while maintaining well-calibrated uncertainty bounds. 

Furthermore, our ablation studies investigated three key variations: first, we examined the impact of fine-tuning all model parameters, denoted as full parameter fine-tuning (FP) in Table~\ref{tab:modelsperffm}; second, we explored replacing the static covariate encoder (SCE, which generates vector C) with an MLP implementation; and third, the performances of conventional methods that employ either MLP or SCE module as input feature compressor, for processing covariates concatenated with target time series. Notably, most optimal performances are achieved without full parameter optimization, yielding significant memory efficiency gains (up to 59\% reduction in memory usage) as in Table~\ref{tab:gpumemusage}.
\begin{figure}[htbp]
    \centering
    \begin{subfigure}[t]{0.32\textwidth}
        \centering
        \includegraphics[width=\textwidth]{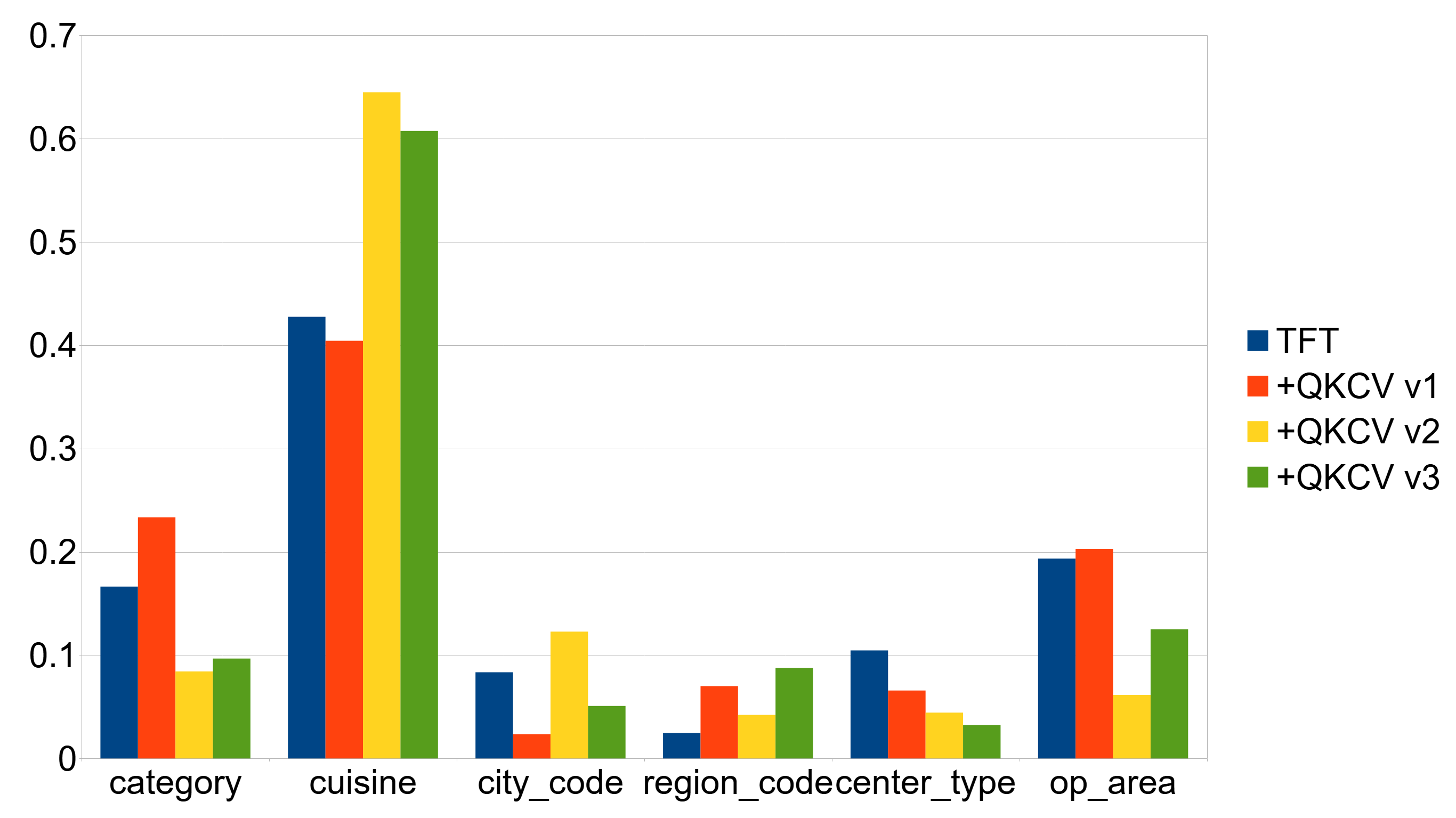}
        \caption{Meal}
        \label{fig:impst:figsub0}
    \end{subfigure}
    \hfill
    \begin{subfigure}[t]{0.32\textwidth}
        \centering
        \includegraphics[width=\textwidth]{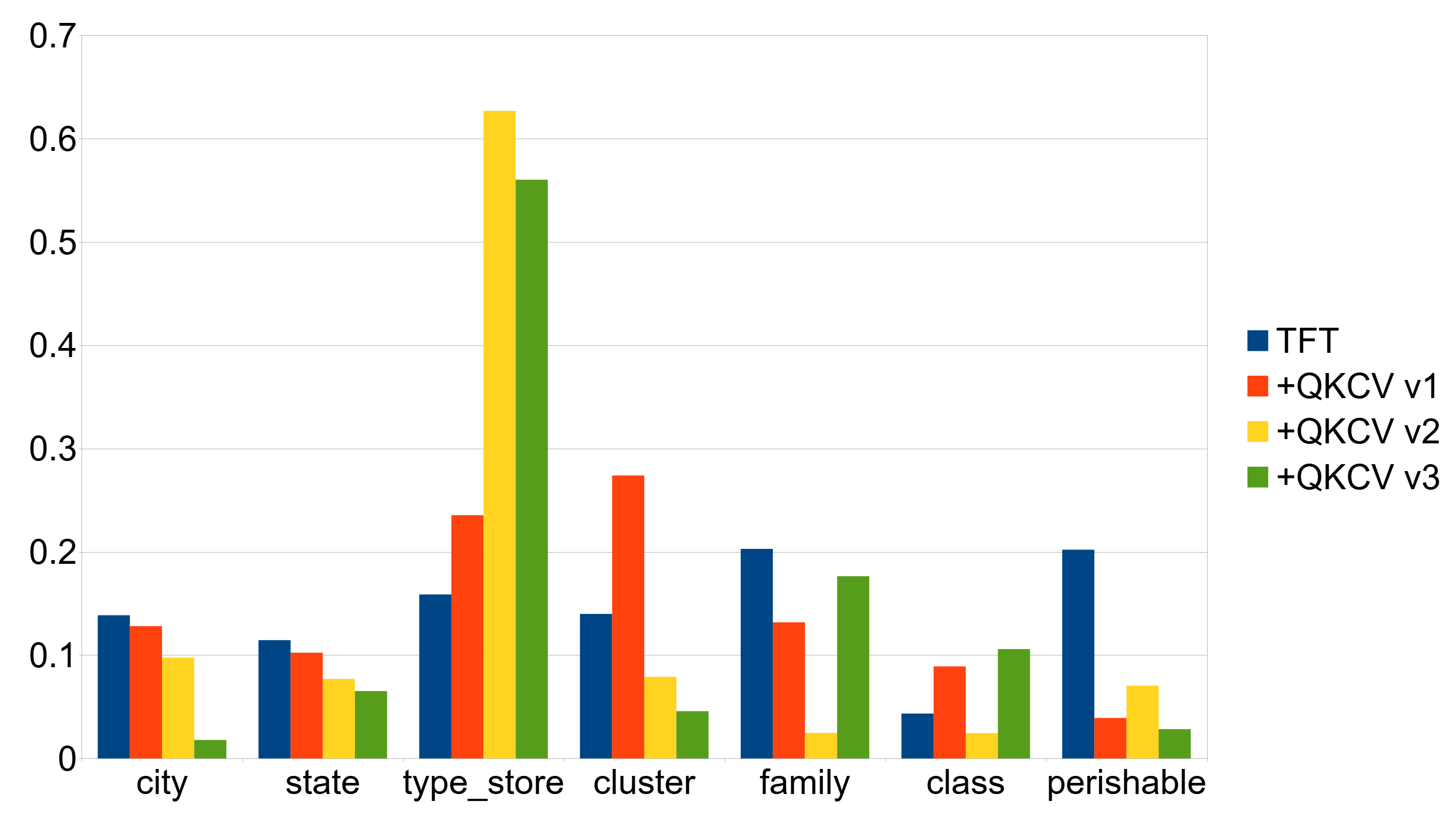}
        \caption{Favorita}
        \label{fig:impst:figsub1}
    \end{subfigure}
    \hfill
    \begin{subfigure}[t]{0.32\textwidth}
        \centering
        \includegraphics[width=\textwidth]{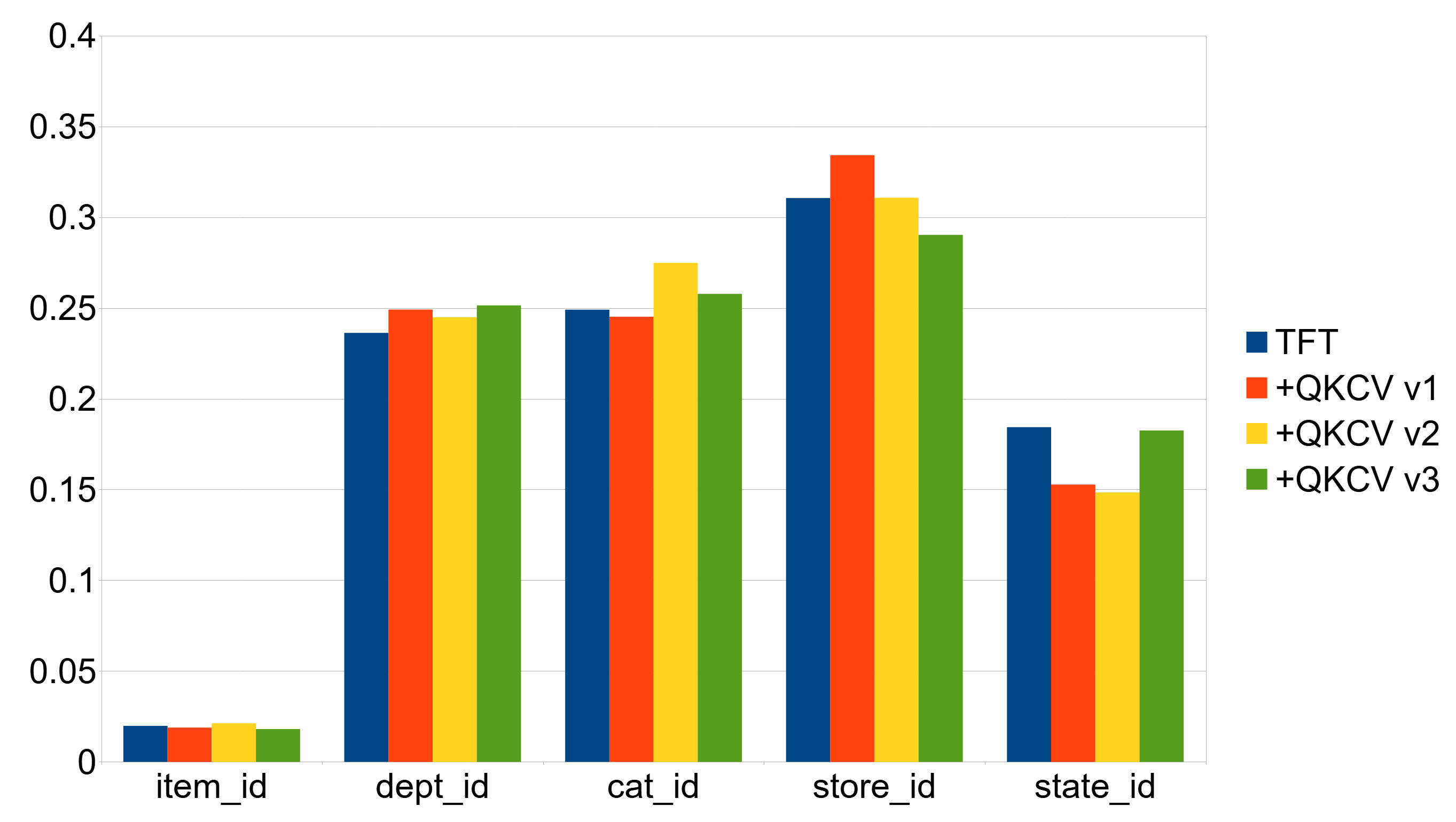}
        \caption{M5}
        \label{fig:impst:figsub2}
    \end{subfigure}
    \caption{Feature importances for static imbeddings in TFT.}
    \label{fig:impst}
\vspace{-0.5em}
\end{figure}

\begin{figure}[htbp]
    \centering
\vspace{-0.5em}
    \begin{subfigure}[t]{0.32\textwidth}
        \centering
        \includegraphics[width=\textwidth]{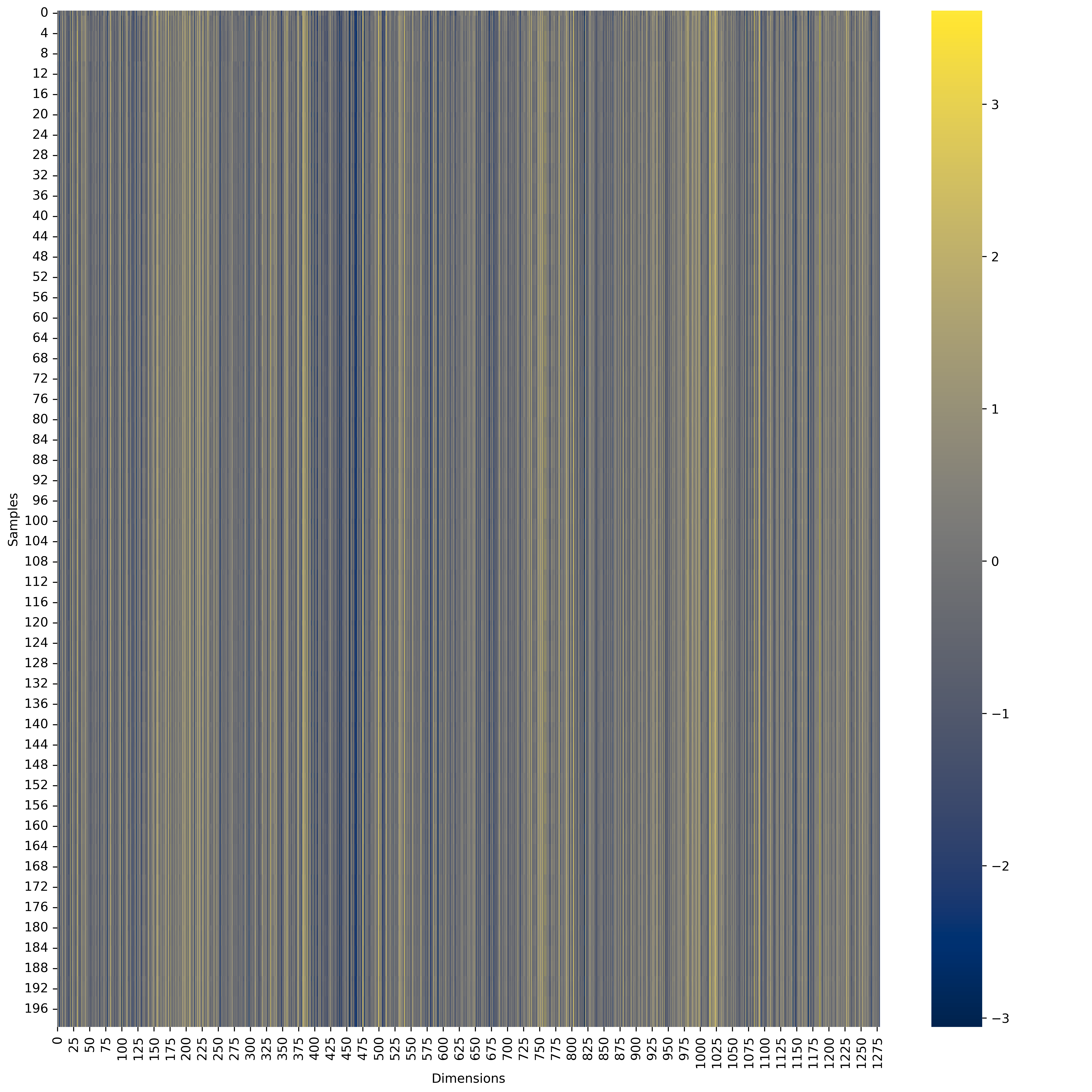}
        \caption{QKCV-v1}
        \label{figmain:attentionscore:figsub01}
    \end{subfigure}
    \hfill
    \begin{subfigure}[t]{0.32\textwidth}
        \centering
        \includegraphics[width=\textwidth]{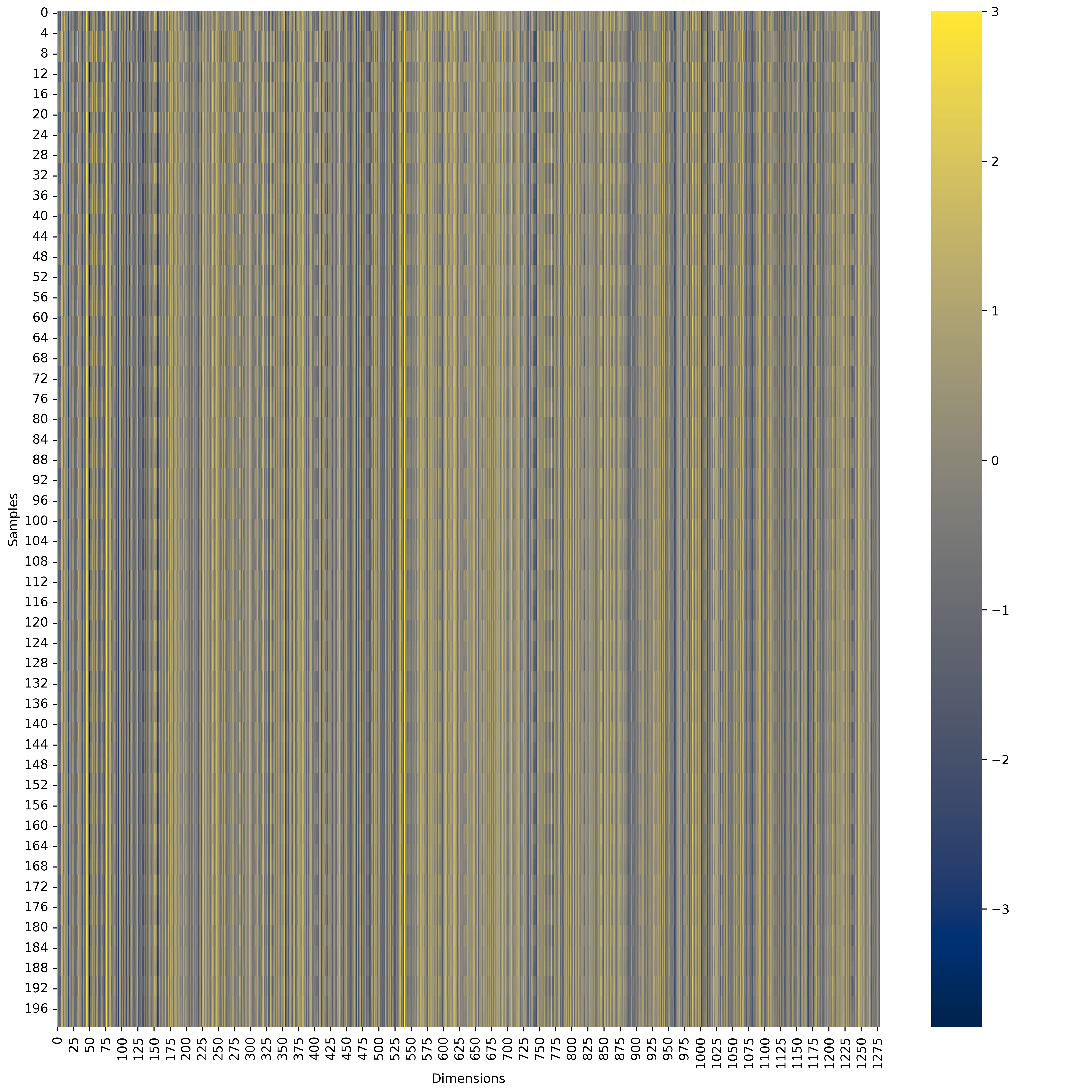}
        \caption{QKCV-v2}
        \label{figmain:attentionscore:figsub02}
    \end{subfigure}
    \begin{subfigure}[t]{0.32\textwidth}
        \centering
        \includegraphics[width=\textwidth]{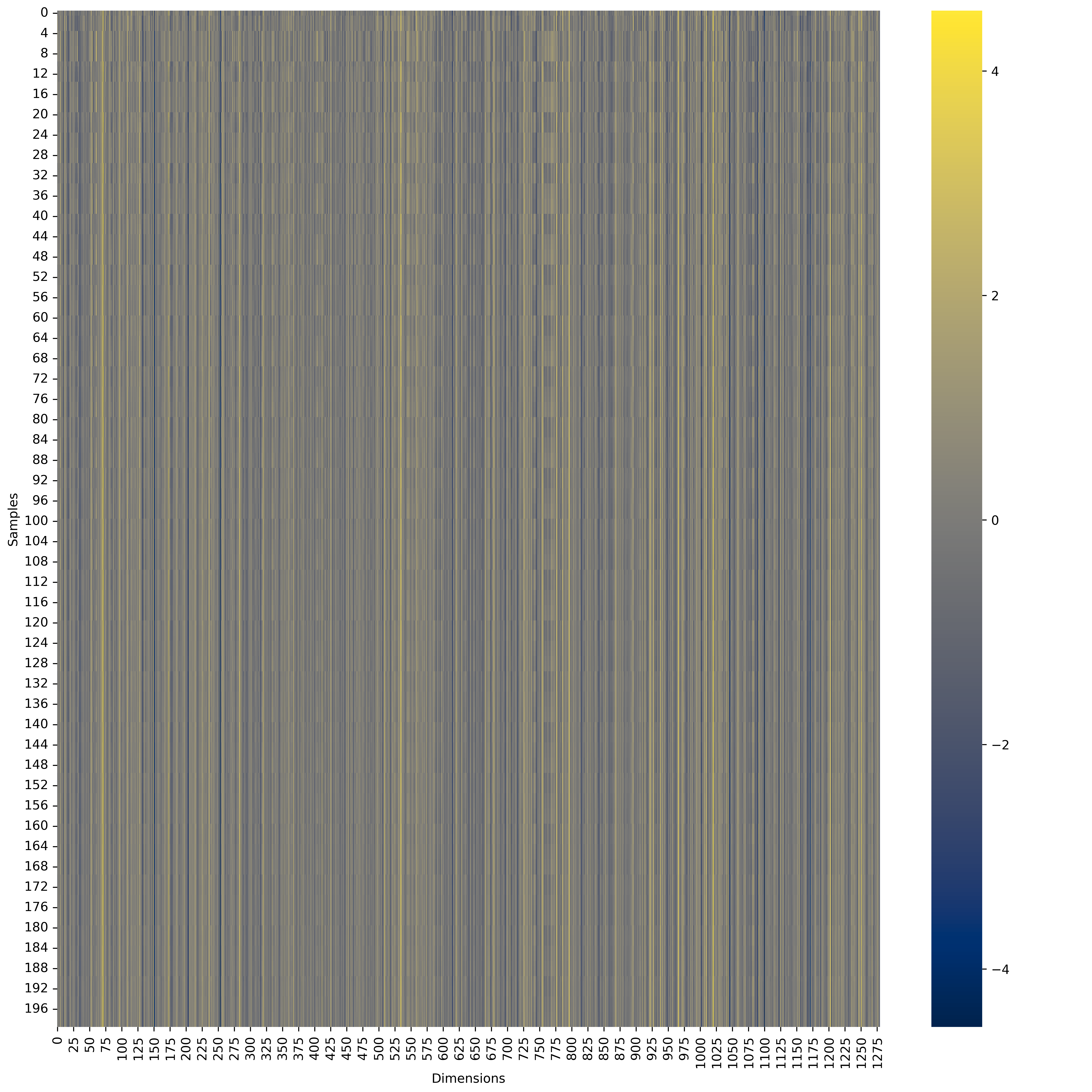}
        \caption{QKCV-v3}
        \label{figmain:attentionscore:figsub03}
    \end{subfigure}
\vspace{0.3cm}
 \begin{subfigure}[t]{0.245\textwidth}
        \centering
        \includegraphics[width=\textwidth]{images/attention\_score\_heat\_M5\_20250512\_QKCV0}
        \caption{no QKCV}
        \label{figmain:attentionscore:figsub08}
    \end{subfigure}
    \hfill
    \begin{subfigure}[t]{0.245\textwidth}
        \centering
        \includegraphics[width=\textwidth]{images/attention\_score\_heat\_M5\_20250512\_QKCV1}
        \caption{QKCV-v1}
        \label{figmain:attentionscore:figsub09}
    \end{subfigure}
    \hfill
    \begin{subfigure}[t]{0.245\textwidth}
        \centering
        \includegraphics[width=\textwidth]{images/attention\_score\_heat\_M5\_20250512\_QKCV2}
        \caption{QKCV-v2}
        \label{figmain:attentionscore:figsub10}
    \end{subfigure}
    \begin{subfigure}[t]{0.245\textwidth}
        \centering
        \includegraphics[width=\textwidth]{images/attention\_score\_heat\_M5\_20250512\_QKCV3}
        \caption{QKCV-v3}
        \label{figmain:attentionscore:figsub11}
    \end{subfigure}
    \caption{Analysis of QKCV attention mechanisms within TimeFM on M5 dataset. (a-c) Learned static embeddings (C) across three QKCV variants, from top 200 samples; (d-f) Associated attention score (QK) matrices of top 100 samples for corresponding subsets. All samples are consistently sorted by entity ID across subfigures. Complete results for all datasets are available in Supplementary Materials.}
    \label{figmain:attentionscore}

\end{figure}

\subsection{Chart analysis}
Given the improved performance on benchmarks, we next demonstrate how categorical embeddings impact the static feature importances. Recognizing that the general interpretability benefits are contingent upon the base model's inherent interpretability capabilities, we conduct an analysis with TFT model from NeuralForecast. Feature importance scores were computed based on the averaged weights of static embeddings derived from the Variable Selection Network (VSN), prior to their input into the GRN, shown in Figure~\ref{fig:impst}, for each dataset and each version of models. The bars in this figure show how the importances of features changed with or without our new attention mechanism. We observe that the distribution of importance values has shifted, with some undergoing significant changes after the application of our new attention mechanism. This could serve as a demonstration of the impact of modifying the attention mechanism on categorical embedding importances, and finally, the model performance. For example, in Figure~\ref{fig:impst:figsub0} meal, the "cuisine" feature importance is increased in the green bar (+QKCV v3), suggesting that the QKCV mechanism considers this feature more important compared to original attention does. It helps to understand which features are crucial for the prediction task, to provides insights into tasks such as feature engineering and feature selection.

To further analyze the impact of QKCV on pre-trained weights, we first visualize the static embeddings (C) from our modified TimeFM architecture using heatmaps. Figures~\ref{figmain:attentionscore}(a)-(c) exhibit distinct vertical patterning across entity samples, with particularly prominent periodicity in Figures~\ref{figmain:attentionscore:figsub02} and~\ref{figmain:attentionscore:figsub03}. This recurrent structure suggests potential categorical similarity or shared characteristic features among these entities. Moreover, comparative analysis between Figure~\ref{figmain:attentionscore:figsub09} to~\ref{figmain:attentionscore:figsub11} and Figure~\ref{figmain:attentionscore:figsub08} demonstrates QKCV's impact on final attention scores of QK. It's evident that the pretrained attention pattern (Figure~\ref{figmain:attentionscore:figsub08}) shows predominantly mean-centered values. In contrast, the QKCV-enhanced versions (Figure~\ref{figmain:attentionscore:figsub09} to Figure~\ref{figmain:attentionscore:figsub11}) exhibit more extreme values, particularly within const dimensions. This indicates that the QKCV mechanism promotes focused attention on specific embedding dimensions, achieving a sparse attention-like effect that likely contributes to the observed performance improvement.

\section{Conclusions}

We propose QKCV Attention, a novel mechanism integrating static embedding C into standard Dot-Product QKV attention, with three variants demonstrating its effectiveness in both lightweight and pre-trained Transformer-based forecasting models. Experiments show that this approach offers: (1) a generalizable framework for incorporating static categorical features into attention mechanisms, (2) efficient fine-tuning of pre-trained foundation models with reduced memory overhead (up to 59\% reduction), and (3) consistent improvements in forecasting accuracy across multiple benchmarks.


\begin{ack}
This research is partly supported by the National Natural Science Foundation of China (72172092), Shanghai Key Laboratory of Brain-Machine Intelligence for Information Behavior (22dz2261100), and the Fundamental Research Funds for the Central Universities (41005067). 

The authors gratefully acknowledge discussions with Mingqi Liu, An Yan, Manav Choudhary, Xiaoying Xiang that contributed to the development of this paper.

\end{ack}

{
\small
\bibliographystyle{unsrt}
\bibliography{biblio}

\begin{thebibliography}{10}

\bibitem{a001}
Zhongyang Han, Jun Zhao, Henry Leung, King~Fai Ma, and Wei Wang.
\newblock A review of deep learning models for time series prediction.
\newblock {\em IEEE Sensors Journal}, 21(6):7833--7848, 2021.

\bibitem{a002}
Zhiping Zeng, Haidong Xiao, and Xinpeng Zhang.
\newblock Self cnn-based time series stream forecasting.
\newblock {\em Electronics Letters}, 52(22):1857--1858, 2016.

\bibitem{rumelhart1986learning}
David~E Rumelhart, Geoffrey~E Hinton, and Ronald~J Williams.
\newblock Learning representations by back-propagating errors.
\newblock {\em nature}, 323(6088):533--536, 1986.

\bibitem{a1989learning}
Ronald~J Williams and David Zipser.
\newblock A learning algorithm for continually running fully recurrent neural
  networks.
\newblock {\em Neural computation}, 1(2):270--280, 1989.

\bibitem{guo2016learning}
Quan Guo, Jia Jia, Guangyao Shen, Lei Zhang, Lianhong Cai, and Zhang Yi.
\newblock Learning robust uniform features for cross-media social data by using
  cross autoencoders.
\newblock {\em Knowledge-Based Systems}, 102:64--75, 2016.

\bibitem{zeng2023financialtimeseriesforecasting}
Zhen Zeng, Rachneet Kaur, Suchetha Siddagangappa, Saba Rahimi, Tucker Balch,
  and Manuela Veloso.
\newblock Financial time series forecasting using cnn and transformer, 2023.

\bibitem{hochreiter1997long}
S~Hochreiter.
\newblock Long short-term memory.
\newblock {\em Neural Computation MIT-Press}, 1997.

\bibitem{hussain2022design}
Hanan Hussain, PS~Tamizharasan, and CS~Rahul.
\newblock Design possibilities and challenges of dnn models: a review on the
  perspective of end devices.
\newblock {\em Artificial Intelligence Review}, pages 1--59, 2022.

\bibitem{huang2019dsanet}
Siteng Huang, Donglin Wang, Xuehan Wu, and Ao~Tang.
\newblock Dsanet: Dual self-attention network for multivariate time series
  forecasting.
\newblock In {\em Proceedings of the 28th ACM international conference on
  information and knowledge management}, pages 2129--2132, 2019.

\bibitem{niu2021review}
Zhaoyang Niu, Guoqiang Zhong, and Hui Yu.
\newblock A review on the attention mechanism of deep learning.
\newblock {\em Neurocomputing}, 452:48--62, 2021.

\bibitem{vaswani2017attention}
A~Vaswani.
\newblock Attention is all you need.
\newblock {\em Advances in Neural Information Processing Systems}, 2017.

\bibitem{lim2021temporal}
Bryan Lim, Sercan~{\"O} Ar{\i}k, Nicolas Loeff, and Tomas Pfister.
\newblock Temporal fusion transformers for interpretable multi-horizon time
  series forecasting.
\newblock {\em International Journal of Forecasting}, 37(4):1748--1764, 2021.

\bibitem{wang2024considering}
Muyao Wang, Wenchao Chen, and Bo~Chen.
\newblock Considering nonstationary within multivariate time series with
  variational hierarchical transformer for forecasting.
\newblock In {\em Proceedings of the AAAI Conference on Artificial
  Intelligence}, volume~38, pages 15563--15570, 2024.

\bibitem{nie2022time}
Yuqi Nie, Nam~H Nguyen, Phanwadee Sinthong, and Jayant Kalagnanam.
\newblock A time series is worth 64 words: Long-term forecasting with
  transformers.
\newblock {\em arXiv preprint arXiv:2211.14730}, 2022.

\bibitem{ouyang2022training}
Long Ouyang, Jeffrey Wu, Xu~Jiang, Diogo Almeida, Carroll Wainwright, Pamela
  Mishkin, Chong Zhang, Sandhini Agarwal, Katarina Slama, Alex Ray, et~al.
\newblock Training language models to follow instructions with human feedback.
\newblock {\em Advances in neural information processing systems},
  35:27730--27744, 2022.

\bibitem{touvron2023llama}
Hugo Touvron, Thibaut Lavril, Gautier Izacard, Xavier Martinet, Marie-Anne
  Lachaux, Timoth{\'e}e Lacroix, Baptiste Rozi{\`e}re, Naman Goyal, Eric
  Hambro, Faisal Azhar, et~al.
\newblock Llama: Open and efficient foundation language models.
\newblock {\em arXiv preprint arXiv:2302.13971}, 2023.

\bibitem{du2021glm}
Zhengxiao Du, Yujie Qian, Xiao Liu, Ming Ding, Jiezhong Qiu, Zhilin Yang, and
  Jie Tang.
\newblock Glm: General language model pretraining with autoregressive blank
  infilling.
\newblock {\em arXiv preprint arXiv:2103.10360}, 2021.

\bibitem{liao2025timegpt}
Wenlong Liao, Shouxiang Wang, Dechang Yang, Zhe Yang, Jiannong Fang, Christian
  Rehtanz, and Fernando Port{\'e}-Agel.
\newblock Timegpt in load forecasting: A large time series model perspective.
\newblock {\em Applied Energy}, 379:124973, 2025.

\bibitem{liu2024generative}
Zhiding Liu, Jiqian Yang, Mingyue Cheng, Yucong Luo, and Zhi Li.
\newblock Generative pretrained hierarchical transformer for time series
  forecasting.
\newblock In {\em Proceedings of the 30th ACM SIGKDD Conference on Knowledge
  Discovery and Data Mining}, pages 2003--2013, 2024.

\bibitem{das2024decoder}
Abhimanyu Das, Weihao Kong, Rajat Sen, and Yichen Zhou.
\newblock A decoder-only foundation model for time-series forecasting.
\newblock In {\em Forty-first International Conference on Machine Learning},
  2024.

\bibitem{tymchuk2022forecasting}
Oleh Tymchuk, Anna Pylypenko, and Maryna Iepik.
\newblock Forecasting of categorical time series using computing with words
  model.
\newblock In {\em IT\&I Workshops}, pages 151--159, 2022.

\bibitem{vanitha2021growth}
SM~Vanitha, Shubhadeep Roy, Neeraj Singh, and Jagdish Singh.
\newblock Growth trend in vegetable production-a time series analysis.
\newblock {\em Journal of Applied Horticulture}, 23(3):294--298, 2021.

\bibitem{oreshkin2019n}
Boris~N Oreshkin, Dmitri Carpov, Nicolas Chapados, and Yoshua Bengio.
\newblock N-beats: Neural basis expansion analysis for interpretable time
  series forecasting.
\newblock {\em arXiv preprint arXiv:1905.10437}, 2019.

\bibitem{challu2023nhits}
Cristian Challu, Kin~G Olivares, Boris~N Oreshkin, Federico~Garza Ramirez,
  Max~Mergenthaler Canseco, and Artur Dubrawski.
\newblock Nhits: Neural hierarchical interpolation for time series forecasting.
\newblock In {\em Proceedings of the AAAI conference on artificial
  intelligence}, volume~37, pages 6989--6997, 2023.

\bibitem{salem2022gated}
Fathi~M Salem and Fathi~M Salem.
\newblock Gated rnn: The gated recurrent unit (gru) rnn.
\newblock {\em Recurrent Neural Networks: From Simple to Gated Architectures},
  pages 85--100, 2022.

\bibitem{graves2012long}
Alex Graves and Alex Graves.
\newblock Long short-term memory.
\newblock {\em Supervised sequence labelling with recurrent neural networks},
  pages 37--45, 2012.

\bibitem{salinas2020deepar}
David Salinas, Valentin Flunkert, Jan Gasthaus, and Tim Januschowski.
\newblock Deepar: Probabilistic forecasting with autoregressive recurrent
  networks.
\newblock {\em International journal of forecasting}, 36(3):1181--1191, 2020.

\bibitem{han2021transformer}
Kai Han, An~Xiao, Enhua Wu, Jianyuan Guo, Chunjing Xu, and Yunhe Wang.
\newblock Transformer in transformer.
\newblock {\em Advances in neural information processing systems},
  34:15908--15919, 2021.

\bibitem{zhou2021informer}
Haoyi Zhou, Shanghang Zhang, Jieqi Peng, Shuai Zhang, Jianxin Li, Hui Xiong,
  and Wancai Zhang.
\newblock Informer: Beyond efficient transformer for long sequence time-series
  forecasting.
\newblock In {\em Proceedings of the AAAI conference on artificial
  intelligence}, volume~35, pages 11106--11115, 2021.

\bibitem{li2019enhancing}
Shiyang Li, Xiaoyong Jin, Yao Xuan, Xiyou Zhou, Wenhu Chen, Yu-Xiang Wang, and
  Xifeng Yan.
\newblock Enhancing the locality and breaking the memory bottleneck of
  transformer on time series forecasting.
\newblock {\em Advances in neural information processing systems}, 32, 2019.

\bibitem{piao2024fredformer}
Xihao Piao, Zheng Chen, Taichi Murayama, Yasuko Matsubara, and Yasushi Sakurai.
\newblock Fredformer: Frequency debiased transformer for time series
  forecasting.
\newblock In {\em Proceedings of the 30th ACM SIGKDD Conference on Knowledge
  Discovery and Data Mining}, pages 2400--2410, 2024.

\bibitem{shchur2023autogluon}
Oleksandr Shchur, Ali~Caner Turkmen, Nick Erickson, Huibin Shen, Alexander
  Shirkov, Tony Hu, and Bernie Wang.
\newblock Autogluon--timeseries: Automl for probabilistic time series
  forecasting.
\newblock In {\em International Conference on Automated Machine Learning},
  pages 9--1. PMLR, 2023.

\bibitem{hollmann2022tabpfn}
Noah Hollmann, Samuel M{\"u}ller, Katharina Eggensperger, and Frank Hutter.
\newblock Tabpfn: A transformer that solves small tabular classification
  problems in a second.
\newblock {\em arXiv preprint arXiv:2207.01848}, 2022.

\bibitem{devalal2018lora}
Shilpa Devalal and A~Karthikeyan.
\newblock Lora technology-an overview.
\newblock In {\em 2018 second international conference on electronics,
  communication and aerospace technology (ICECA)}, pages 284--290. IEEE, 2018.

\bibitem{dettmers2023qlora}
Tim Dettmers, Artidoro Pagnoni, Ari Holtzman, and Luke Zettlemoyer.
\newblock Qlora: Efficient finetuning of quantized llms.
\newblock {\em Advances in neural information processing systems},
  36:10088--10115, 2023.

\bibitem{kopiczko2023vera}
Dawid~J Kopiczko, Tijmen Blankevoort, and Yuki~M Asano.
\newblock Vera: Vector-based random matrix adaptation.
\newblock {\em arXiv preprint arXiv:2310.11454}, 2023.

\bibitem{bhardwaj2024rapid}
Kartikeya Bhardwaj, Nilesh~Prasad Pandey, Sweta Priyadarshi, Viswanath
  Ganapathy, Rafael Esteves, Shreya Kadambi, Shubhankar Borse, Paul Whatmough,
  Risheek Garrepalli, Mart Van~Baalen, et~al.
\newblock Rapid switching and multi-adapter fusion via sparse high rank
  adapters.
\newblock {\em arXiv preprint arXiv:2407.16712}, 2024.

\bibitem{bergstra2011algorithms}
James Bergstra, R{\'e}mi Bardenet, Yoshua Bengio, and Bal{\'a}zs K{\'e}gl.
\newblock Algorithms for hyper-parameter optimization.
\newblock {\em Advances in neural information processing systems}, 24, 2011.

\bibitem{olivares2022libraryneuralforecast}
Kin~G. Olivares, Cristian Challú, Federico Garza, Max~Mergenthaler Canseco,
  and Artur Dubrawski.
\newblock {NeuralForecast}: User friendly state-of-the-art neural forecasting
  models.
\newblock {PyCon} Salt Lake City, Utah, US 2022, 2022.

\end{thebibliography}
}


\appendix

\section{Technical Appendices and Supplementary Material}

\subsection{Datasets}

\paragraph{\textbf{Meal}} The meal demand forecasting dataset originates from a meal delivery company that operates fulfillment centers in multiple cities for food delivery. Each entity is uniquely identified by a combination of center-meal ID pairs, totaling 3,589 entities. This dataset comprises static metadata regarding the nature of meals and delivery centers, making it suitable for demonstrating forecasting performances. We examined the dynamic features in this dataset and found that they may encompass future information like checkout price and promotion, and therefore dynamic features in this dataset were not utilized in our experiments. For validation, we utilize training samples spanning from week 20 to week 125, consisting of 336,433 records, with the validation set comprises 32,820 records from week 126 to week 135, following the training set. For performance evaluation, the test set includes 32,821 records from week 136 to week 145, with 369,253 records from week 20 to week 135 as training set. Data is sampled on a weekly basis, with missing weeks filled using the last available observation. Our objective is to forecast the order number for the subsequent 10 weeks.

\paragraph{\textbf{Favorita}} The retail dataset from Favorita encompasses the unit sales of items sold at various Favorita stores located in Ecuador and has been utilized as a benchmark for TFT model. This dataset includes comprehensive static metadata and observed time-varying inputs. The feature processing procedure is similar to \cite{lim2021temporal}. Each entity is uniquely identified by a combination of product number and store number pairs, totaling over 143k entities. For training, we utilize samples spanning from 2015-01-01 to 2015-11-30, consisting of approximately 25 million records. The validation set comprises 2.8 million records from 2015-12-01 to 2015-12-30, following the training set. The test set includes 2.9 million records from the 30 days afterwards. Data is resampled at regular daily intervals, with any missing days filled using the last available observation via the forward fill (ffill) method. Notably, not all entities have a value on the first day; therefore, we search up to the beginning of the dataset to locate the last observation. If no such data is available, such as when some entities' first sale occurs after January 1, 2015, we fill the missing sales at the beginning with a placeholder value of 0. We do not perform the log-transformation described in \cite{lim2021temporal} due to the presence of negative unit sales values, likely resulting from returned sales.

\paragraph{\textbf{M5}} This forecasting dataset comes from the fifth Makridakis Competition, containing hierarchical sales data from Walmart, the world’s largest company by revenue. The dataset contains 59m records with 30k entities. The entity identifer is generated from item\_id and store\_id. In our experiment, 57m samples between 2011-01-29 and 2016-03-27(d1-d1885)  are used as training set, 853k samples between 2016-03-28 and 2016-04-24(d1886-d1913)  are validation set, while last 853k samples between 2016-04-25 and 2016-05-22(d1914 - d1941) are test set. We use attributes of each item and store, including departments, catalogs of items, and locations of stores, as the static catalogical features. The target daily sales quantity has a range between 0 and 763. Forecasting was performed for 28-day horizion. 

\subsection{Hyperparameters search ranges}

The full search ranges for common hyperparameters in model tunning are detailed as below:

 \begin{itemize}
  \item \textbf{input size}  - 2*horizon, 3*horizon
  \item \textbf{hidden size}  - 32, 64, 128, 256
  \item \textbf{n head}  - 1, 2, 4
  \item \textbf{learning rate}  - 0.01, 0.001, 0.0001
  \item \textbf{dropout}  - 0.1, 0.2, 0.3
  \item \textbf{max steps}  -  500, 1000, 3000, 5000
  \item \textbf{batch size}  - 64, 128, 256, 512
  \item \textbf{windows batch size}  - 256, 512, 1024
\end{itemize}

\subsection{Additional figures}

\begin{figure}[htbp]
    \centering
    \begin{subfigure}[t]{0.32\textwidth}
        \centering
        \includegraphics[width=\textwidth]{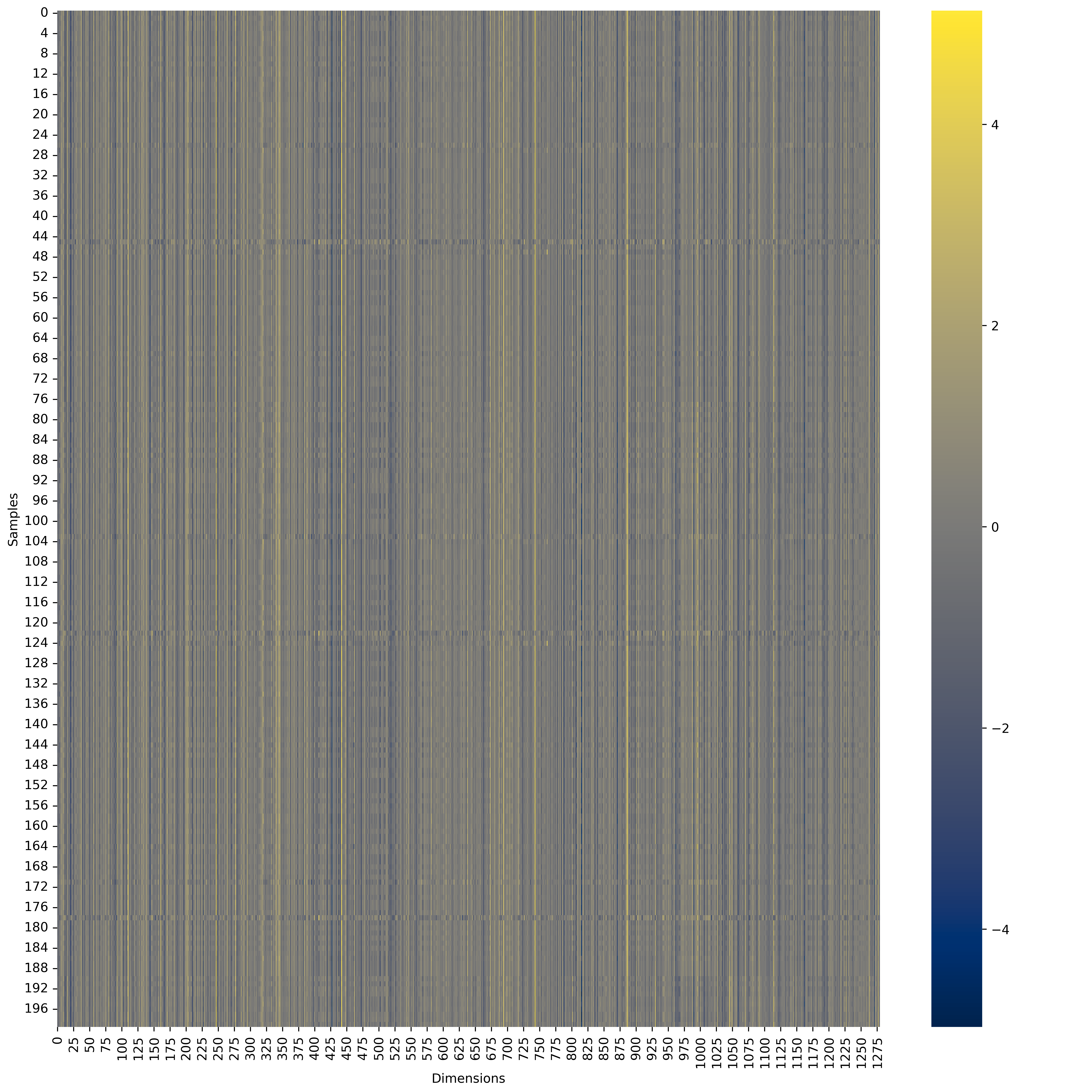}
        \caption{Meal QKCV-v1}
        \label{figmain:sec:200:figsub01}
    \end{subfigure}
    \hfill
    \begin{subfigure}[t]{0.32\textwidth}
        \centering
        \includegraphics[width=\textwidth]{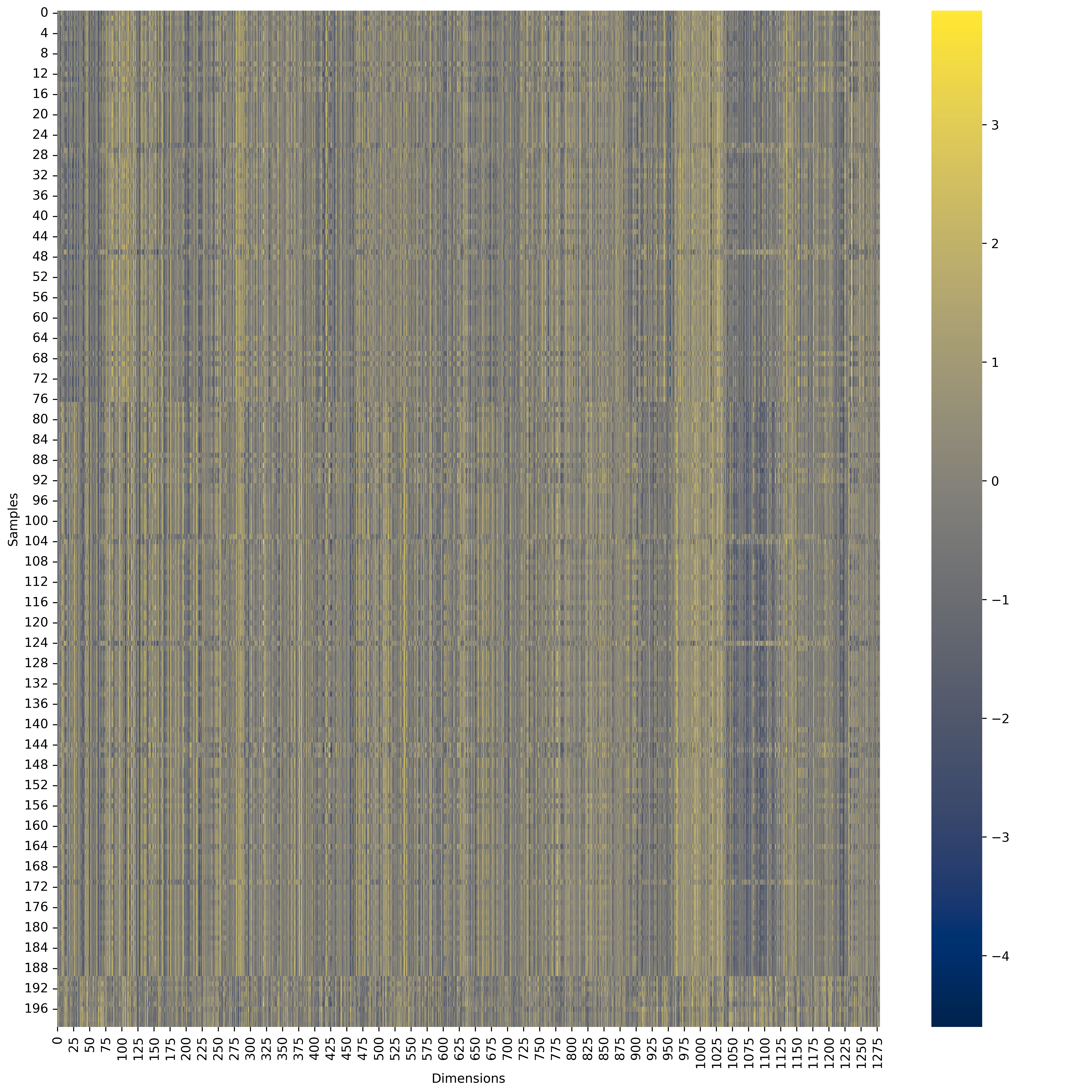}
        \caption{Meal QKCV-v2}
        \label{figmain:sec:200:figsub02}
    \end{subfigure}
    \begin{subfigure}[t]{0.32\textwidth}
        \centering
        \includegraphics[width=\textwidth]{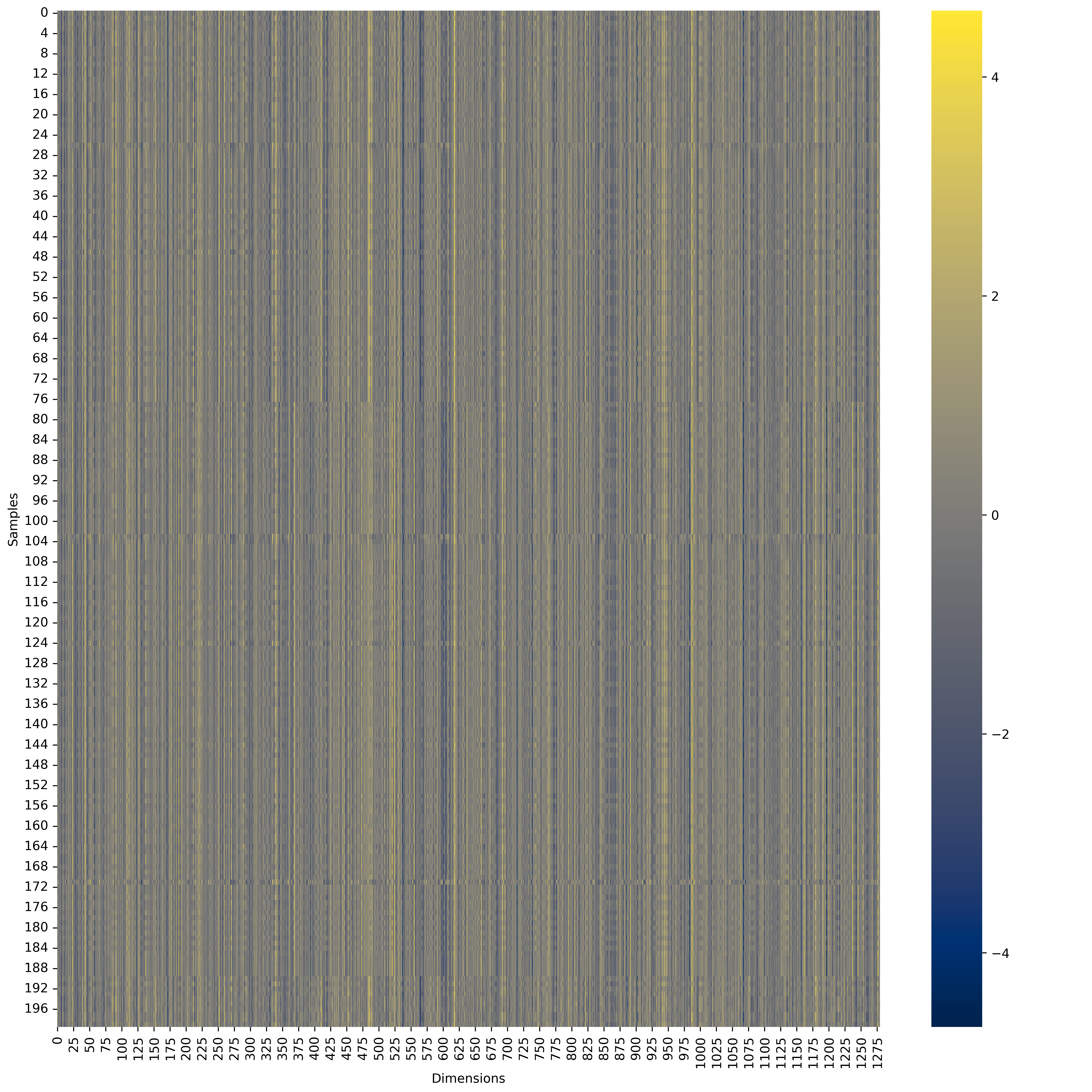}
        \caption{Meal QKCV-v3}
        \label{figmain:sec:200:figsub03}
    \end{subfigure}
\vspace{0.3cm}
\begin{subfigure}[t]{0.32\textwidth}
        \centering
        \includegraphics[width=\textwidth]{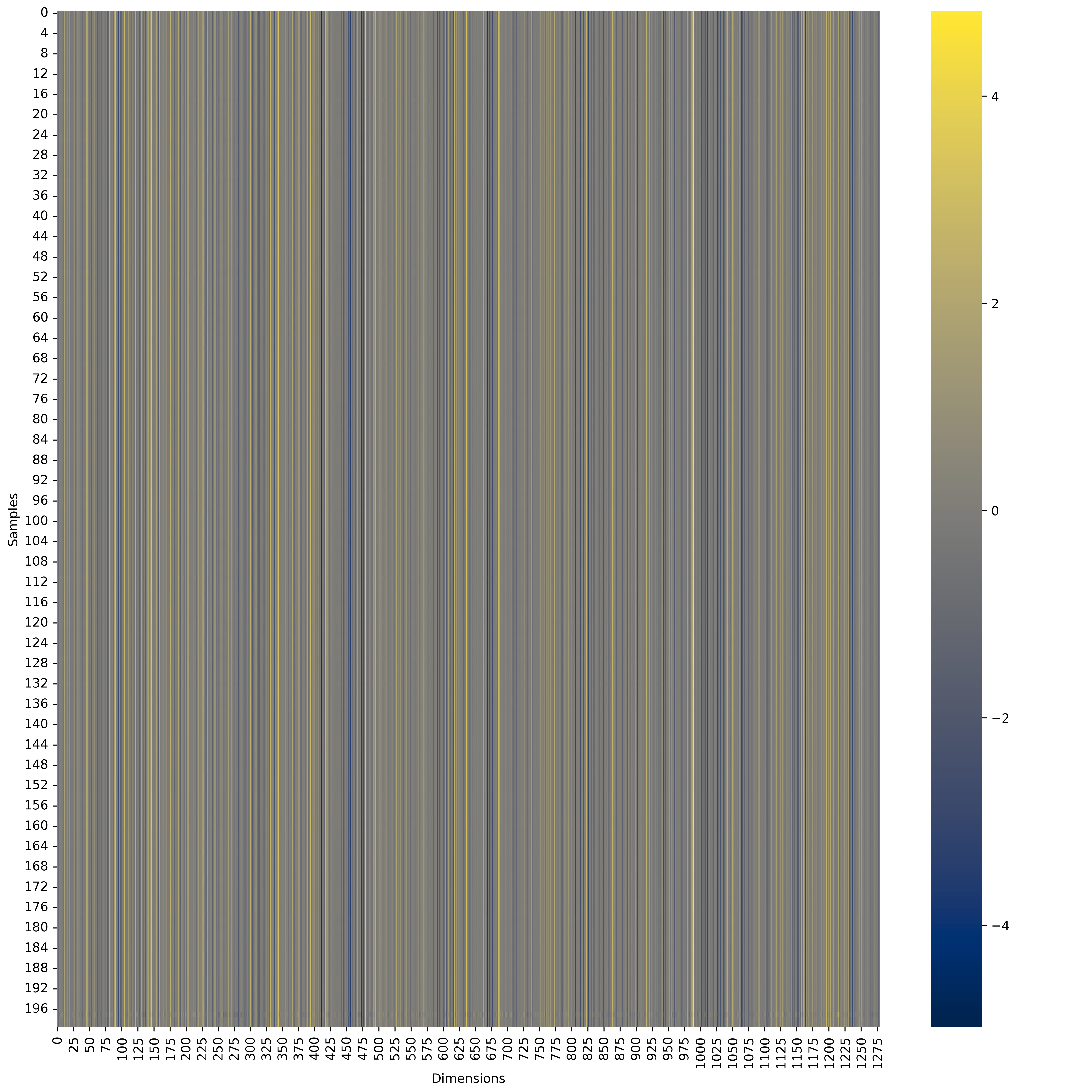}
        \caption{Favorita QKCV-v1}
        \label{figmain:sec:200:figsub11}
    \end{subfigure}
    \hfill
    \begin{subfigure}[t]{0.32\textwidth}
        \centering
        \includegraphics[width=\textwidth]{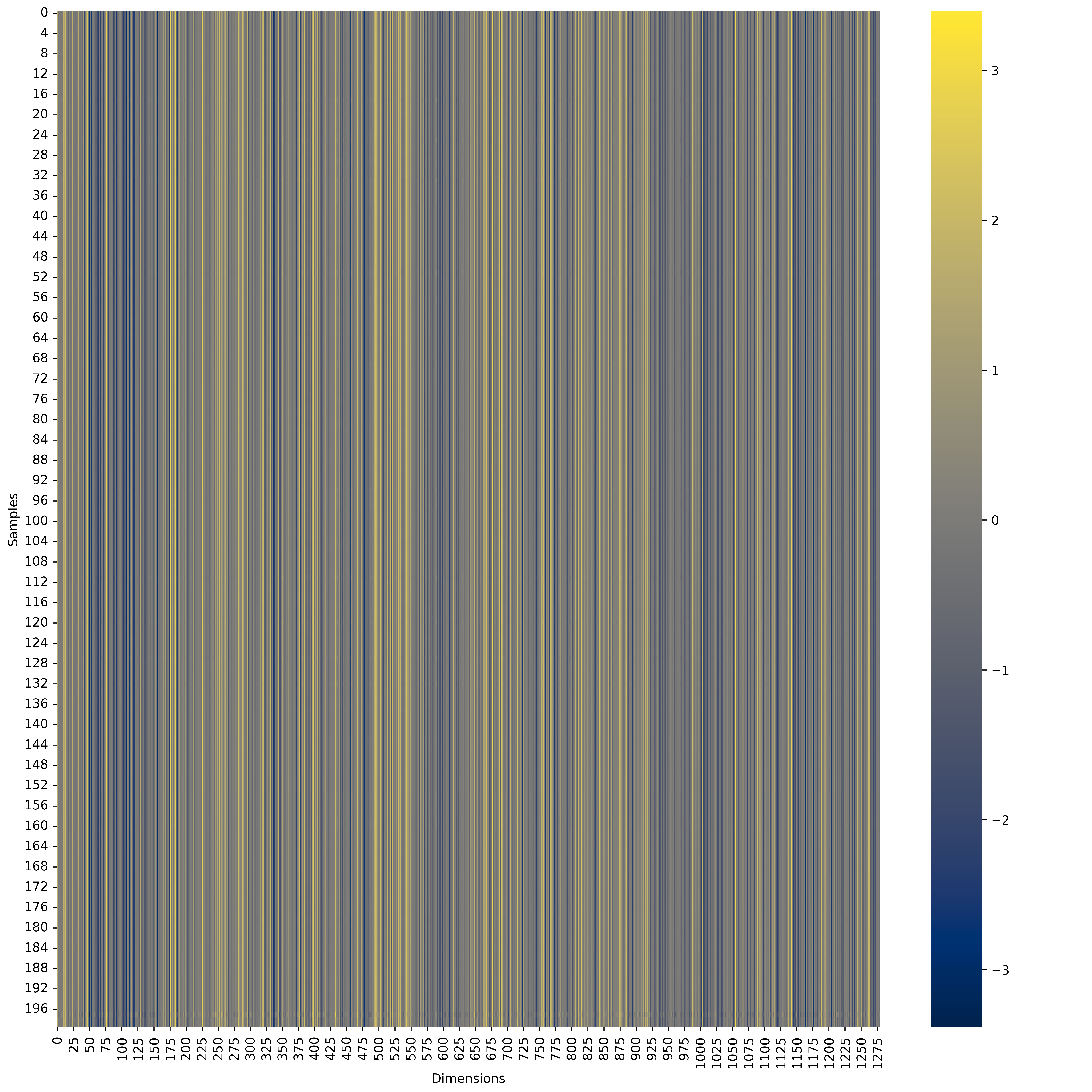}
        \caption{Favorita QKCV-v2}
        \label{figmain:sec:200:figsub12}
    \end{subfigure}
    \begin{subfigure}[t]{0.32\textwidth}
        \centering
        \includegraphics[width=\textwidth]{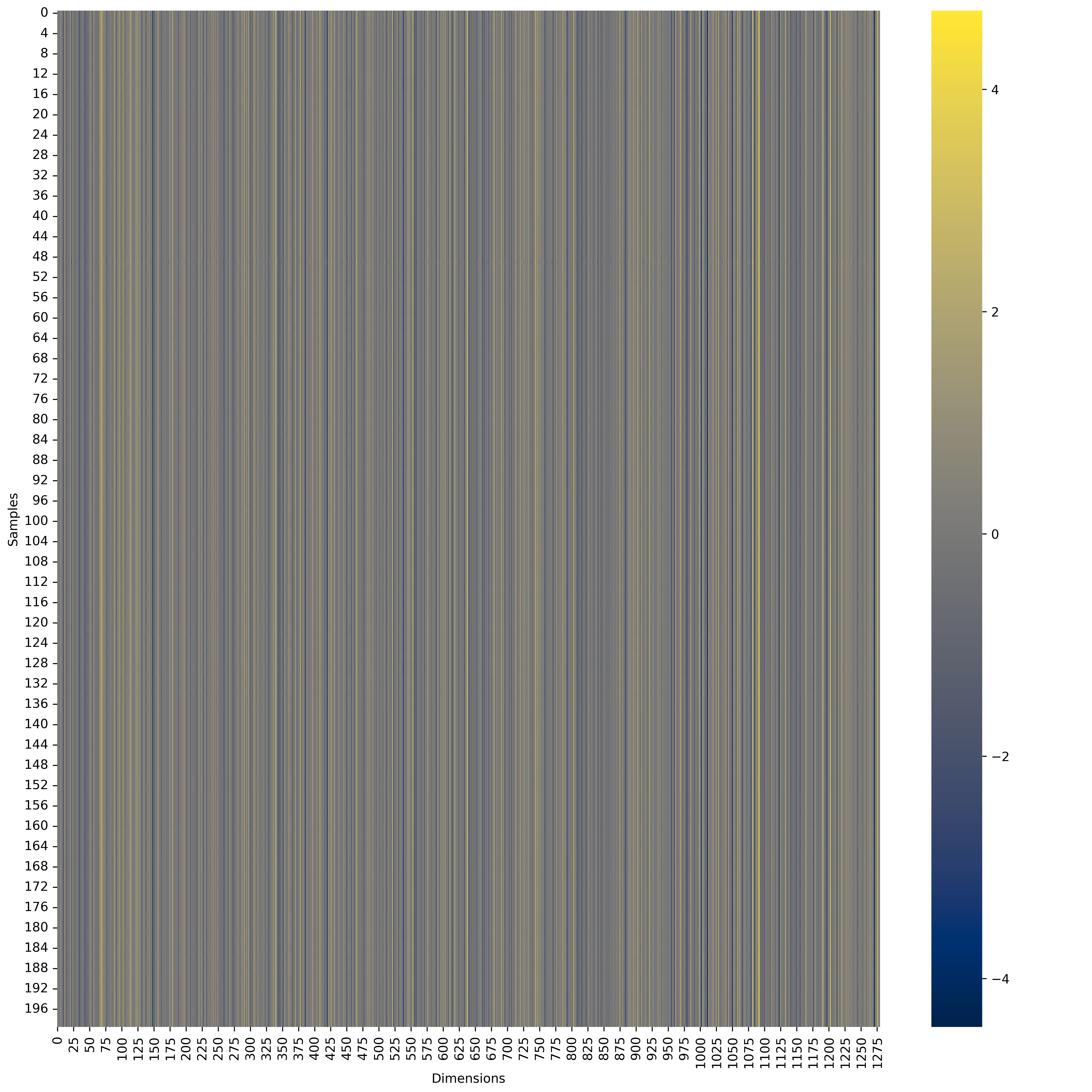}
        \caption{Favorita QKCV-v3}
        \label{figmain:sec:200:figsub13}
    \end{subfigure}
\vspace{0.3cm}
\begin{subfigure}[t]{0.32\textwidth}
        \centering
        \includegraphics[width=\textwidth]{images/caheatM520250514QKCV1.png}
        \caption{M5 QKCV-v1}
        \label{figmain:sec:200:figsub21}
    \end{subfigure}
    \hfill
    \begin{subfigure}[t]{0.32\textwidth}
        \centering
        \includegraphics[width=\textwidth]{images/caheatM520250514QKCV2.png}
        \caption{M5 QKCV-v2}
        \label{figmain:sec:200:figsub22}
    \end{subfigure}
    \begin{subfigure}[t]{0.32\textwidth}
        \centering
        \includegraphics[width=\textwidth]{images/caheatM520250514QKCV3.png}
        \caption{M5 QKCV-v3}
        \label{figmain:sec:200:figsub23}
    \end{subfigure}

    \caption{Static embedding C of top 200 test samples for each dataset.}
    \label{figmain:sec:200}
\end{figure}

\begin{figure}[htbp]
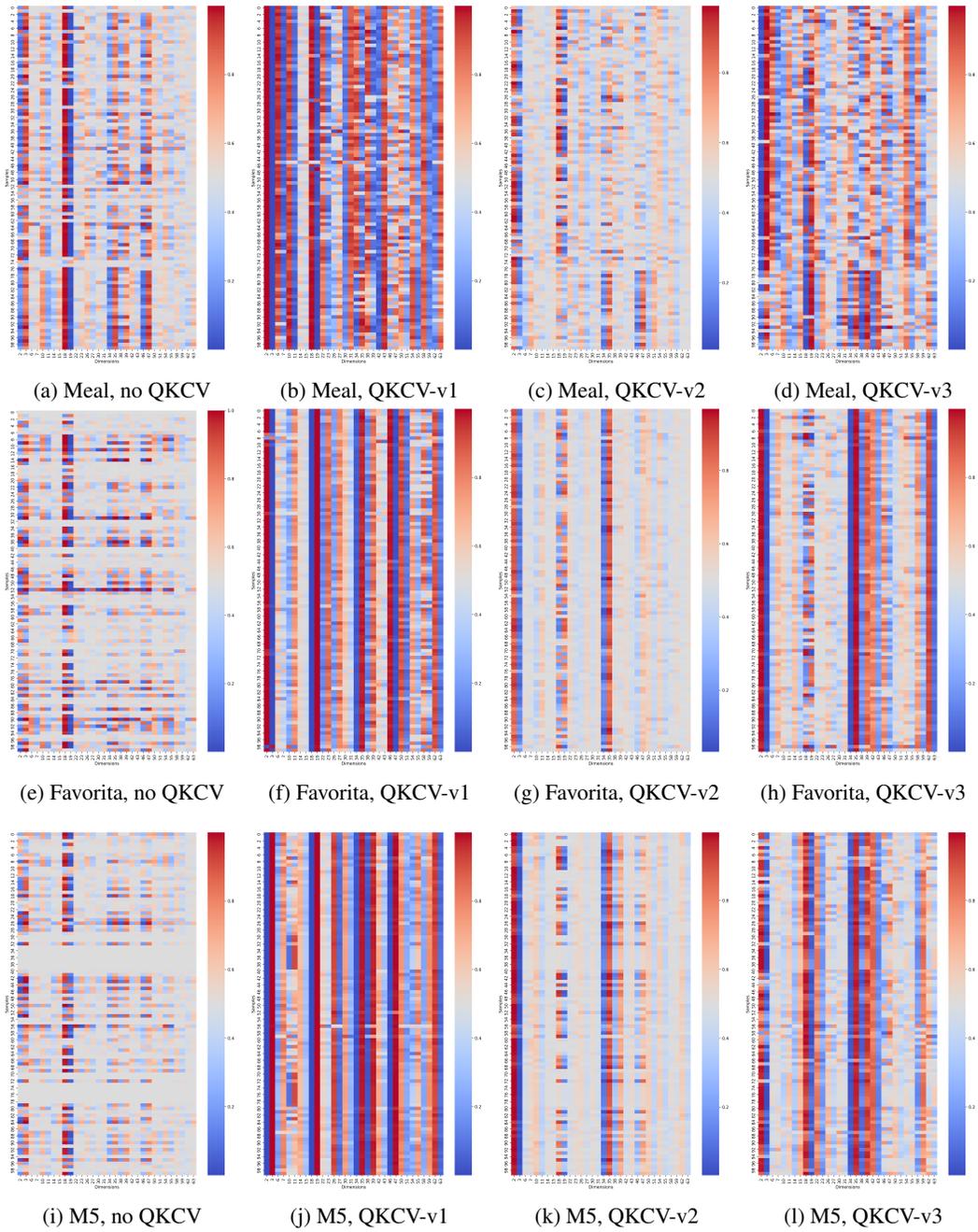

    \centering
    \begin{subfigure}[t]{0.245\textwidth}
        \centering
        \includegraphics[width=\textwidth]{images/attention\_score\_heat\_Meal\_20250512\_QKCV0}
        \caption{Meal, no QKCV}
        \label{figmain:attentionscore:all:figsub00}
    \end{subfigure}
    \hfill
    \begin{subfigure}[t]{0.245\textwidth}
        \centering
        \includegraphics[width=\textwidth]{images/attention\_score\_heat\_Meal\_20250512\_QKCV1}
        \caption{Meal, QKCV-v1}
        \label{figmain:attentionscore:all:figsub01}
    \end{subfigure}
    \hfill
    \begin{subfigure}[t]{0.245\textwidth}
        \centering
        \includegraphics[width=\textwidth]{images/attention\_score\_heat\_Meal\_20250512\_QKCV2}
        \caption{Meal, QKCV-v2}
        \label{figmain:attentionscore:all:figsub02}
    \end{subfigure}
    \begin{subfigure}[t]{0.245\textwidth}
        \centering
        \includegraphics[width=\textwidth]{images/attention\_score\_heat\_Meal\_20250512\_QKCV3}
        \caption{Meal, QKCV-v3}
        \label{figmain:attentionscore:all:figsub03}
    \end{subfigure}
	\vspace{0.3cm}
 \begin{subfigure}[t]{0.245\textwidth}
        \centering
        \includegraphics[width=\textwidth]{images/attention\_score\_heat\_Favorita\_20250512\_QKCV0}
        \caption{Favorita, no QKCV}
        \label{figmain:attentionscore:all:figsub04}
    \end{subfigure}
    \hfill
    \begin{subfigure}[t]{0.245\textwidth}
        \centering
        \includegraphics[width=\textwidth]{images/attention\_score\_heat\_Favorita\_20250512\_QKCV1}
        \caption{Favorita, QKCV-v1}
        \label{figmain:attentionscore:all:figsub05}
    \end{subfigure}
    \hfill
    \begin{subfigure}[t]{0.245\textwidth}
        \centering
        \includegraphics[width=\textwidth]{images/attention\_score\_heat\_Favorita\_20250512\_QKCV2}
        \caption{Favorita, QKCV-v2}
        \label{figmain:attentionscore:all:figsub06}
    \end{subfigure}
    \begin{subfigure}[t]{0.245\textwidth}
        \centering
        \includegraphics[width=\textwidth]{images/attention\_score\_heat\_Favorita\_20250512\_QKCV3}
        \caption{Favorita, QKCV-v3}
        \label{figmain:attentionscore:all:figsub07}
    \end{subfigure}
\vspace{0.3cm}
 \begin{subfigure}[t]{0.245\textwidth}
        \centering
        \includegraphics[width=\textwidth]{images/attention\_score\_heat\_M5\_20250512\_QKCV0}
        \caption{M5, no QKCV}
        \label{figmain:attentionscore:all:figsub08}
    \end{subfigure}
    \hfill
    \begin{subfigure}[t]{0.245\textwidth}
        \centering
        \includegraphics[width=\textwidth]{images/attention\_score\_heat\_M5\_20250512\_QKCV1}
        \caption{M5, QKCV-v1}
        \label{figmain:attentionscore:all:figsub09}
    \end{subfigure}
    \hfill
    \begin{subfigure}[t]{0.245\textwidth}
        \centering
        \includegraphics[width=\textwidth]{images/attention\_score\_heat\_M5\_20250512\_QKCV2}
        \caption{M5, QKCV-v2}
        \label{figmain:attentionscore:all:figsub10}
    \end{subfigure}
    \begin{subfigure}[t]{0.245\textwidth}
        \centering
        \includegraphics[width=\textwidth]{images/attention\_score\_heat\_M5\_20250512\_QKCV3}
        \caption{M5, QKCV-v3}
        \label{figmain:attentionscore:all:figsub11}
    \end{subfigure}
    \caption{Attention scores of top 100 test samples for each dataset.}
    \label{figmain:attentionscore:all}
\end{figure}


\newpage
\section*{NeurIPS Paper Checklist}

\begin{enumerate}

\item {\bf Claims}
    \item[] Question: Do the main claims made in the abstract and introduction accurately reflect the paper's contributions and scope?
    \item[] Answer: \answerYes{} 
    \item[] Justification: The paper explicitly delineates its contributions and scope through rigorous methodological analysis and experimental validation.

\item {\bf Limitations}
    \item[] Question: Does the paper discuss the limitations of the work performed by the authors?
    \item[] Answer: \answerNA{} 
    \item[] Justification: {At this stage of research, we have not identified significant limitations that would affect the reported findings within our defined scope of work.}

\item {\bf Theory assumptions and proofs}
    \item[] Question: For each theoretical result, does the paper provide the full set of assumptions and a complete (and correct) proof?
    \item[] Answer: \answerNA{} 
    \item[] Justification: {Our paper focuses primarily on empirical methodology and experimental validation.}

    \item {\bf Experimental result reproducibility}
    \item[] Question: Does the paper fully disclose all the information needed to reproduce the main experimental results of the paper to the extent that it affects the main claims and/or conclusions of the paper (regardless of whether the code and data are provided or not)?
    \item[] Answer: \answerYes{} 
    \item[] Justification: {Yes, the paper provides complete methodological details in Sections 4 to reproduce the core experiments.}

\item {\bf Open access to data and code}
    \item[] Question: Does the paper provide open access to the data and code, with sufficient instructions to faithfully reproduce the main experimental results, as described in supplemental material?
    \item[] Answer: \answerYes{} 
    \item[] Justification: {Yes, our paper provides full open access to both code and data through public repo.}

\item {\bf Experimental setting/details}
    \item[] Question: Does the paper specify all the training and test details (e.g., data splits, hyperparameters, how they were chosen, type of optimizer, etc.) necessary to understand the results?
    \item[] Answer: \answerYes{} 
    \item[] Justification: {Yes, the paper provides a comprehensive and meticulous account of all the training and test details essential for understanding the results, in Section 4.1, 4.2 and Supplementary Materials.}

\item {\bf Experiment statistical significance}
    \item[] Question: Does the paper report error bars suitably and correctly defined or other appropriate information about the statistical significance of the experiments?
    \item[] Answer: \answerNo{} 
    \item[] Justification: {The reported results are computed using deterministic evaluation metrics, which exhibit no measurement variability by design. To ensure full reproducibility, all experiments were conducted with fixed random seeds (seed=42 for all trials).}

\item {\bf Experiments compute resources}
    \item[] Question: For each experiment, does the paper provide sufficient information on the computer resources (type of compute workers, memory, time of execution) needed to reproduce the experiments?
    \item[] Answer: \answerYes{} 
    \item[] Justification: {GPU utilization metrics are explicitly documented in Table~\ref{tab:gpumemusage}, while other critical system specifications (e.g., CPU, RAM, storage) are omitted due to their negligible impact on the reported results in our experimental paradigm.}
    
\item {\bf Code of ethics}
    \item[] Question: Does the research conducted in the paper conform, in every respect, with the NeurIPS Code of Ethics \url{https://neurips.cc/public/EthicsGuidelines}?
    \item[] Answer: \answerYes{} 
    \item[] Justification: {Yes, this research fully complies with the NeurIPS Code of Ethics in all aspects.}

\item {\bf Broader impacts}
    \item[] Question: Does the paper discuss both potential positive societal impacts and negative societal impacts of the work performed?
    \item[] Answer: \answerNA{} 
    \item[] Justification: {No, while our paper thoroughly addresses technical contributions and experimental validation, it does not explicitly analyze societal impacts due to the fundamental nature of this research.}
    
\item {\bf Safeguards}
    \item[] Question: Does the paper describe safeguards that have been put in place for responsible release of data or models that have a high risk for misuse (e.g., pretrained language models, image generators, or scraped datasets)?
    \item[] Answer: \answerNA{} 
    \item[] Justification: {Our technical contributions and experimental artifacts present little societal risk.}

\item {\bf Licenses for existing assets}
    \item[] Question: Are the creators or original owners of assets (e.g., code, data, models), used in the paper, properly credited and are the license and terms of use explicitly mentioned and properly respected?
    \item[] Answer: \answerYes{} 
    \item[] Justification: {Yes, all third-party assets are properly credited and compliant with their original licenses.}

\item {\bf New assets}
    \item[] Question: Are new assets introduced in the paper well documented and is the documentation provided alongside the assets?
    \item[] Answer: \answerYes{} 
    \item[] Justification: {Yes, the new assets introduced in the paper are well-documented, and the corresponding documentation is provided alongside the assets.}

\item {\bf Crowdsourcing and research with human subjects}
    \item[] Question: For crowdsourcing experiments and research with human subjects, does the paper include the full text of instructions given to participants and screenshots, if applicable, as well as details about compensation (if any)? 
    \item[] Answer: \answerNA{} 
    \item[] Justification: {The paper does not involve crowdsourcing.}

\item {\bf Institutional review board (IRB) approvals or equivalent for research with human subjects}
    \item[] Question: Does the paper describe potential risks incurred by study participants, whether such risks were disclosed to the subjects, and whether Institutional Review Board (IRB) approvals (or an equivalent approval/review based on the requirements of your country or institution) were obtained?
    \item[] Answer: \answerNA{} 
    \item[] Justification: {The paper does not involve crowdsourcing.}

\item {\bf Declaration of LLM usage}
    \item[] Question: Does the paper describe the usage of LLMs if it is an important, original, or non-standard component of the core methods in this research? Note that if the LLM is used only for writing, editing, or formatting purposes and does not impact the core methodology, scientific rigorousness, or originality of the research, declaration is not required.
    \item[] Answer: \answerNA{} 
    \item[] Justification: {No, the paper does not discuss LLM usage as they are not employed in our core methodology.}

\end{enumerate}

\end{document}